\documentclass[]{fairmeta}
\usepackage{amsthm,amsmath,amssymb}
\usepackage{graphicx}
\usepackage{natbib}
\usepackage{booktabs}
\usepackage{tabularx}
\usepackage{makecell}

\usepackage{algorithm}
\usepackage{algpseudocode} 
\usepackage{tcolorbox}
\usepackage{enumitem}
\usepackage{multirow}
\usepackage{booktabs}

\usepackage{subcaption}
\usepackage{booktabs}
\usepackage[misc]{ifsym} 
\newtheorem{theorem}{Theorem}[]

\newtheorem{remark1}[theorem]{Remark}

\newcommand{\modelname}{Full-4D}

\title{\modelname{}: Generating Full-Scope  4D Scene  from a Single-View Video}

\author[1,2]{Tingxi Chen\textsuperscript{*}}
\author[1,2]{Ke Hao\textsuperscript{*}}
\author[2]{Yabo Chen\textsuperscript{$\dagger$}}
\author[1]{Zhengxue Cheng}
\author[1]{Rong Xie}
\author[1]{Li Song\textsuperscript{\Letter}}
\author[2]{Haibin Huang}
\author[2]{Chi Zhang}
\author[2]{Xuelong Li\textsuperscript{\textbf{\Letter}}}

\affiliation[1]{Shanghai Jiao Tong University}
\affiliation[2]{Institute of Artificial Intelligence, China Telecom (TeleAI)}

\contribution{\textsuperscript{*}Equal Contributions}
\contribution{\textsuperscript{$\dagger$}Project Leader}

\begin{document}

\abstract{Generating 4D scenes from a single-view video is inherently ill-posed: a single viewpoint lacks the information needed to recover a complete, dynamic scene with full coverage. Existing methods are typically limited to monocular videos, simple 3D effects, or only small viewpoint perturbations around the original viewpoint, falling short of true 4D generation. 
Meanwhile, the lack of large-scale datasets capturing full-scope 4D scenes with synchronized multi-view videos further hinders progress in this direction.
We propose a novel single-view video-to-4D framework that casts full-scope 4D generation as a multi-view video synthesis followed by optimization-based 4D reconstruction from the generated views. To instantiate this formulation end-to-end, we make three key contributions.
First, we introduce Real-MV-4D, a large-scale dataset of synchronized multi-view videos captured in diverse real-world environments to provide the 4D supervision. Second, we train a multi-view video diffusion model driven by a novel fused time(T)-view(V) attention mechanism that directly embeds geometric reprojection priors and explicit camera conditioning into its view-time interactions. Unlike basic feature fusion, this direct binding strictly aligns the generation process with physical 3D priors to produce a dense, synchronized T$\times $V video grid. Third, rather than relying on non-interactive and inconsistent 2D video interpolations, we lift the synthesized multi-view videos into an explicit 4D representation (\textit{i.e.} 4DGS), regularized by a Flow Matching Distillation loss that exploits the multi-view prior to improve novel-view rendering. Extensive experiments demonstrate that our method outperforms existing approaches in both visual fidelity and geometric consistency, enabling full-scope 4D scene generation from single-view videos. See more at https://ccxi1008.github.io/Full-4D/.}

\maketitle
\begin{figure}
    \centering
    \includegraphics[width=1\linewidth]{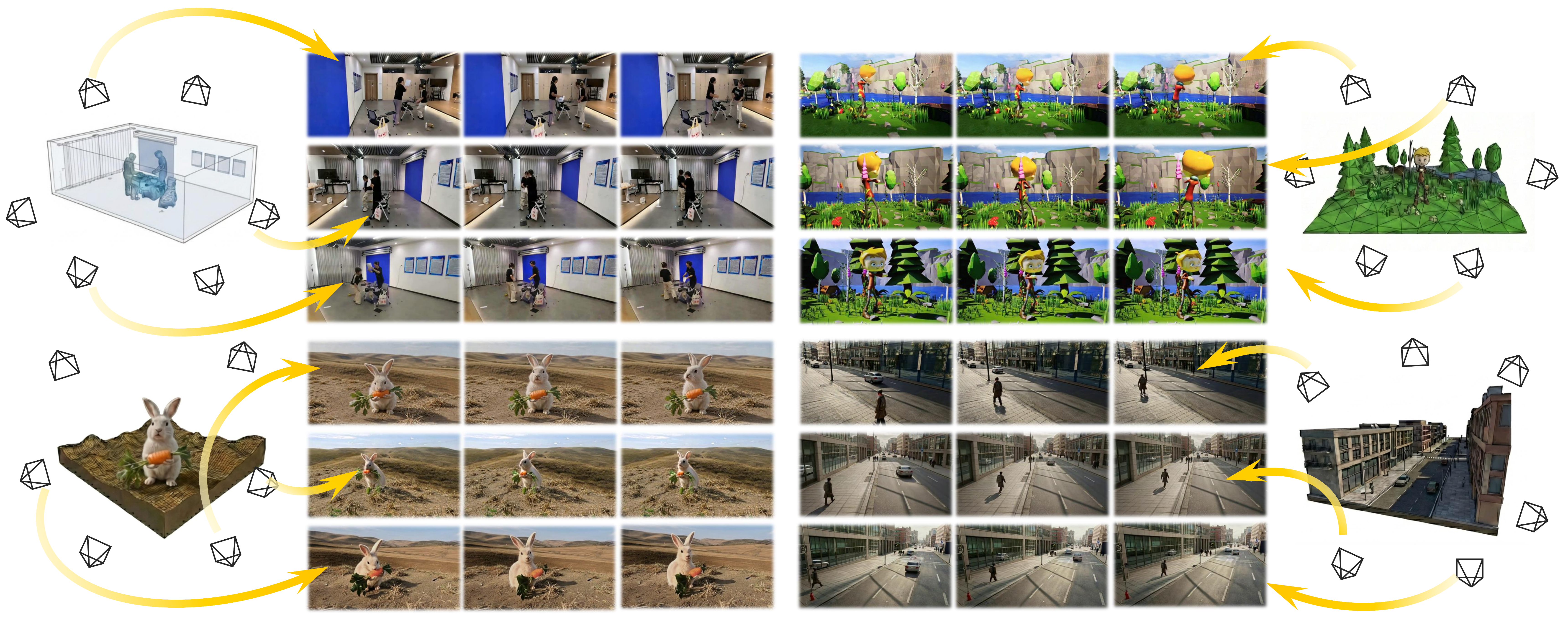}
    \vspace{-5mm}
    \caption{Given a single-view input video, our Full-4D generates a synchronized multi-view 4D video grid over large angular ranges and lifts into 4D scene representation. The rendered results demonstrate strong generalization across diverse scenes and styles.}
    \label{fig:teaser}
\end{figure}

\section{Introduction}

The ability to create immersive, dynamic 4D scenes is central to next-generation applications in virtual reality, film production, and the metaverse~\cite{guo2025ctrl, wang2024drivedreamer, liang2024dreamitate}. Recent advances in diffusion models have enabled high-quality generation of images, videos, and 3D assets~\cite{guo2023animatediff, gao2025seedance, fu2025learning, brooks2024video, ali2025world, shi2023mvdream, wu2024direct3d, chen2023single, he2024gvgen,zhang2026symphomotion,zhang2026telephysics,xiang2025macro,chen2024liftimage3d,chen2025teleworld,chen2024cascadezero123,huang2025zero,eccv_domainfusion,NEURIPS2025_DDPTM,chen2026dp}. However, extending these capabilities to 4D scene generation from a monocular video remains challenging, as it requires consistent modeling across both time and viewpoints~\cite{fu2026plenoptic, van2026anyview, cao2026freeorbit4d, chen2026beyond, yang2026neoverse, miao2025advances, zhang20244diffusion,li20244k4dgen,liang2024diffusion4d,zhao2024genxd,zheng2024unified,li2024dreammesh4d,song2025worldforge,rahamim2024bringing}.

Existing  methods generally fall into two categories: reconstruction-based and generation-based approaches. Many existing techniques only focus on object-centric 4D creation and are rarely applied to complex full scenes due to scarcity of paired real-world data~\cite{4d-survey,park2025zero4d,objcet-sv4d,object-puppeteer,object-track,objet-fb4dd}. 
For 4D scenes, approaches that reconstruct from a monocular video are inherently limited by the observed camera trajectory and cannot recover unseen viewpoints, restricting the angular coverage of the resulting scene~\cite{vdpm,hu2025vggt4d,karhade2025any4d}. Some recent works attempt to alleviate this limitation through camera-controlled video synthesis~\cite{recammaster,syncammaster,trajectorycrafter,jeong2025reangle} or multi-view video generation~\cite{4realvideo,4realvideov2,liu2025free4d,dimensionx,zhao2024genxd,objcet-sv4d}. 
However, a fundamental gap exists between these tasks and true 4D generation. Camera-controlled baselines primarily synthesize moving monocular videos where time and viewpoint shift simultaneously (coupled), lacking the ability to produce multi-view outputs for a single timestamp. When forced into a decoupled 4D setting to generate a dense time-view grid, these methods suffer from severe structural distortions and inconsistencies, especially at large angles. Consequently, current methods struggle to ensure both \textbf{temporal continuity} across time $T$ and \textbf{wide coverage} across viewpoints $V$. The problem is further compounded by the lack of large-scale datasets that provide paired supervision between monocular videos and synchronized multi-view observations, which are essential for learning consistent video-to-4D generation.

To address these challenges, we first introduce Real-MV-4D, a large-scale dataset of synchronized multi-view videos captured from diverse real-world scenes, differing from existing resources that typically provide monocular videos, limited-view captures, synthetic or object-centric data~\cite{data-eitke2023objaverse,data-ling2024dl3dv,data-nan2024openvid,data-zhou2018stereo,recammaster,syncammaster}. This design provides the missing training signal for learning consistent scene structure across both time and viewpoints, which is essential for video-to-4D modeling.

Building upon this dataset, we develop a unified framework for reconstructing a 4D scene from a single-view input video through joint time–view modeling. Rather than treating temporal dynamics and multi-view synthesis as separate processes, we formulate 4D generation as learning a coupled distribution over a synchronized $T \times V$ video grid. In practice, the entire grid is generated jointly using a single multi-view video diffusion model. Crucially, to transcend basic feature fusion, we introduce a novel fused time–view attention mechanism that directly embeds geometric reprojection priors and explicit camera conditioning into its view-time interactions. By directly binding these physical 3D priors within the attention computation, our model strictly aligns the generation process with the multi-view geometry, ensuring coherent motion over time and consistent structure across arbitrary viewpoints.

While the generated T$ \times$ V video grid provides dense observations, directly relying on it for novel view synthesis is insufficient for true 4D generation. Utilizing 2D interpolation or repeated diffusion for intermediate views lacks spatio-temporal consistency and is both computationally expensive and non-interactive. To address this, the generated multi-view videos are subsequently lifted into an explicit 4D Gaussian representation. This fits all views jointly into one shared 4D model, enabling continuous viewpoint changes and interactive free-view rendering. 
To further improve reconstruction quality, we introduce a Flow Matching Distillation (FMD) loss that incorporates the generative prior of the pretrained video diffusion model during optimization. This regularization encourages the reconstructed scene to remain consistent with the learned video distribution, improving cross-view consistency and stabilizing geometry in regions that are weakly constrained by the input video.

Together, these components enable coherent 4D scene reconstruction from a single-view video while maintaining consistency across both time and viewpoints.

Our contributions are summarized as follows:
\begin{itemize}
    \item We introduce Real-MV-4D, a large-scale real-world dataset of synchronized multi-view videos captured with 6 cameras, covering 2k+ scenes and 6k+ subjects, providing paired data across both time and views.
    \item  We propose a multi-view video diffusion model driven by a fused sparse time–view attention mechanism that directly embeds geometric reprojection and camera priors. This jointly generates a synchronized T $\times$V video grid in a single pass, achieving true multi-view consistency per timestamp (decoupled time and viewpoints) rather than moving monocular videos.
    \item  We integrate the generative model with a 4D Gaussian Splatting pipeline, regularized by a flow matching distillation loss, to lift videos into explicit 4D scene representations.

\end{itemize}

\section{Related Works}
\subsection{4D Video Generation.} Recent advances in diffusion-based video synthesis have significantly improved the quality and scalability of video generation models ~\cite{videodiffusionmodels,svd,harvey2022flexible,sd3,flux,yang2024cogvideox,wan,peebles2023scalable-dit,ho2020denoising}. However, most existing video foundation models focus on generating monocular videos and do not explicitly model multi-view geometry.

One line of work explores camera-controlled video generation~~\cite{recammaster,syncammaster,trajectorycrafter}, where user-specified camera trajectories are used to synthesize novel views and produce dynamic shots such as dolly zooms or mild orbital motions. However, these methods typically generate only a single monocular video where time and viewpoint shift simultaneously (\textit{i.e.}, coupled dynamics without spatial-temporal decoupling). Moreover, the resulting viewpoint variations are often limited to small perturbations around the input perspective—insufficient for recovering a complete 4D scene.

More recently, some approaches aim to construct 4D representations by generating synchronized video grids across both time and viewpoints $(T\times V)$~\cite{4realvideo,4realvideov2,cat4d,dimensionx,liu2025free4d}. However, many of these approaches obtain different viewpoints through repeated single-view generation rather than jointly modeling multi-view outputs, which often leads to inconsistencies when viewpoint changes become large. As a result, while some methods can successfully output a 4D representation, their high-quality synthesis is practically restricted to small angular variations, struggling to achieve full-scope 4D coverage~\cite{chen20254dnex,mi2025one4d,zhang2025joint,ren2025gen3c}. Furthermore, most methods that achieve wide viewpoint coverage are trained on synthetic, object-centric datasets ~\cite{dreamgaussian4d,4dfy,animate124,4real,objcet-sv4d,object-puppeteer,objet-fb4dd,object-track}, which limits their applicability to complex real-world scenes.

\subsection{4D Reconstruction}
Recovering 4D representations from 2D observations is a classic ill-posed problem. Recent works have explored feed-forward reconstruction that direct reconstruction from monocular inputs~\cite{vdpm,hu2025vggt4d,depthanything,karhade2025any4d}, which often heavily rely on estimated depth maps. However, they are inherently limited to synthesizing geometrically consistent content for unobserved angles, and their performance degrades significantly with sparse input views. Alternatively, optimization-based approaches~\cite{dreamgaussian4d,4real}, based on  Dynamic NeRF~\cite{nerf} or 4D Gaussian Splatting~\cite{4dgs,wu2025swift4d}, iteratively fit a representation to the input views. Although more computationally intensive, they offer superior fidelity and consistency. In this work, we adopt 4D Gaussian Splatting as our representation. Our optimization is guided by high-quality synchronized multi-view videos generated from our model and is further enhanced by leveraging the pretrained video diffusion prior through a distillation loss.

\section{Method}
\begin{figure}[t]
    \centering
    \includegraphics[width=1\linewidth]{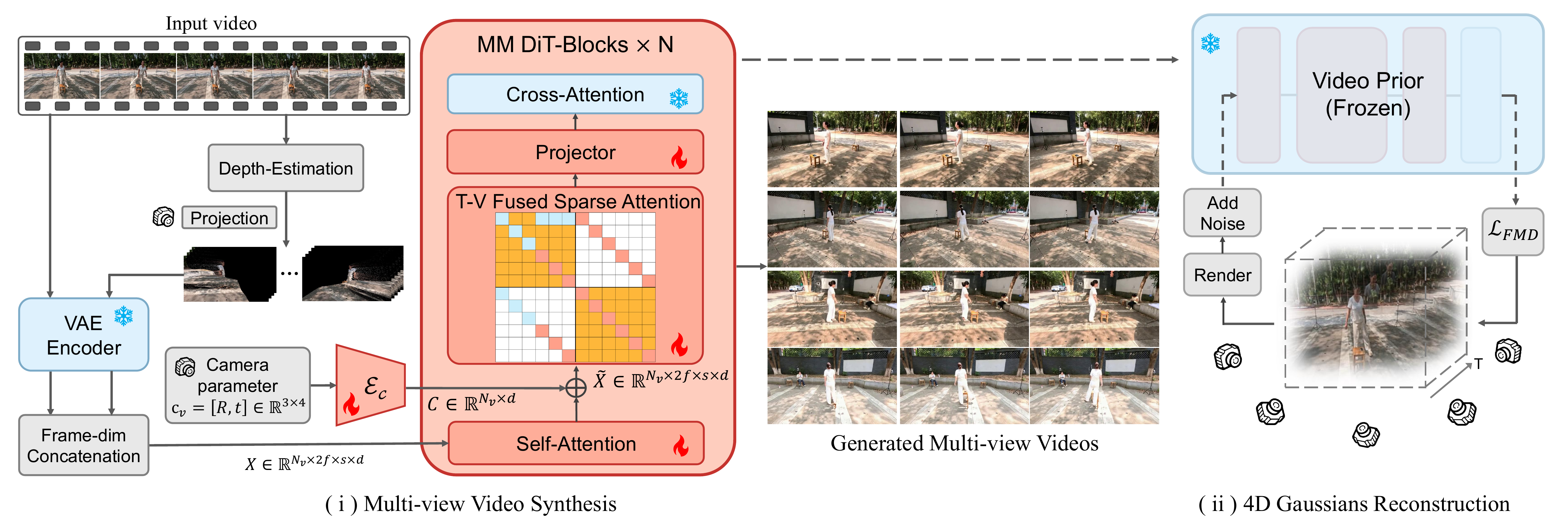}
    \caption{\textbf{Overview of our Full-4D framework.} Given a single-view input video, we first generate synchronized multi-view videos across \(N_v\) target viewpoints at once using our diffusion model, which incorporates spatial-aware projection conditioning and a fused view-time  sparse attention mechanism to ensure  consistency.  These generated videos are then fed into a 4D Gaussian Splatting pipeline. A Flow Matching Distillation (FMD) loss regularizes the optimization by leveraging the pretrained video diffusion prior, filling occluded regions and enhancing consistency. }
    \vspace{-2mm}
    \label{fig:pipeline}
\end{figure}

To generate full-scope 4D scene, our framework consists of (i)multi-view video synthesis via a proposed 4D diffusion model in Sec.\ref{sec:video generation} (ii) Gaussian reconstruction that lifts the generated videos into 4D Gaussian ellipsoids in Sec. \ref{sec:4dgs}, as shown in Fig.\ref{fig:pipeline}.

\subsection{Preliminary}
\textbf{DiT-based denoising model.} Our method builds upon a latent video diffusion framework based on the Diffusion Transformer (DiT). Input videos are first encoded into a latent space using a 3D Variational Auto-Encoder (VAE), and denoised by the model composed of multiple DiT blocks. Each block contains a 3D self-attention module to model spatial–temporal dependencies within each sample, followed by a cross-attention layer for conditioning signals such as text embeddings.

The generative model adopts the Rectified Flow framework for the noise schedule and denoising process, which defines the forward process as a straight-line path between the data distribution and a standard Gaussian prior. Specifically, given a latent sample $z_0 \sim p_0$ and noise $\epsilon \sim \mathcal{N}(0, I)$, the forward interpolation is defined as:
\begin{equation}
    z_t = (1 - t) z_0 + t \epsilon, \quad t \in [0, 1]
\end{equation}
where $t$ denotes the continuous time step. The denoising process is governed by an ordinary differential equation (ODE):

\begin{equation}
    dz_t = v_{\Theta}(z_t, t) \, dt
\end{equation}
where the velocity field $v_{\Theta}$ is parameterized by the DiT with learnable parameters $\Theta$. To train the model, we minimize a Conditional Flow Matching (CFM) loss:

\begin{equation}\label{equation:training loss}
    \mathcal{L}_{\mathrm{CFM}} = \mathbb{E}_{t, p_t(z | \epsilon), p(\epsilon)} \| v_{\Theta}(z_t, t) - u_t(z_0 | \epsilon) \|_2^2
\end{equation}
where $u_t(z_0 | \epsilon)$ is the target vector field derived from the interpolation path. During inference, we solve the ODE using Euler discretization with step size $\Delta t$ over the
timestep interval at [0, 1], starting from $t = 1$:

\begin{equation}
    z_{t} = z_{t-1} - v_{\Theta}(z_{t-1}, t) \cdot \Delta t
\end{equation}

While widely adopted, this framework generates only a single video per inference. Naively extending it to multiple viewpoints by generating each view independently leads to cross-view geometric inconsistencies and temporal misalignment. We therefore extend it to jointly generate a synchronized multi-view video grid, which provides the supervision required for 4D lifting.
\subsection{Multi-view 4D Video Generation}
\label{sec:video generation}
Given an input video $V_s \in \mathbb{R}^{f \times 3 \times h \times w}$ captured from a reference viewpoint 
, our goal is to generate a \textbf{grid} of multi-view videos $\{V_v\}_{v=1}^{N_v}=\{I_{v,t}\}$, where $v \in \{1,\dots,N_v\}$ indexes viewpoints and $t \in \{1,\dots,f\}$ indexes time steps. Each row corresponds to a fixed-view video, and the first row is the source video $V_s$ itself. For each target viewpoint $v$ represented by $\mathrm{cam}_v$, we aim to synthesize a video $V_v \in \mathbb{R}^{f \times 3 \times h \times w}$ that is temporally coherent and geometrically consistent with the source.

\begin{figure}[t]
    \centering
    \includegraphics[width=1\linewidth]{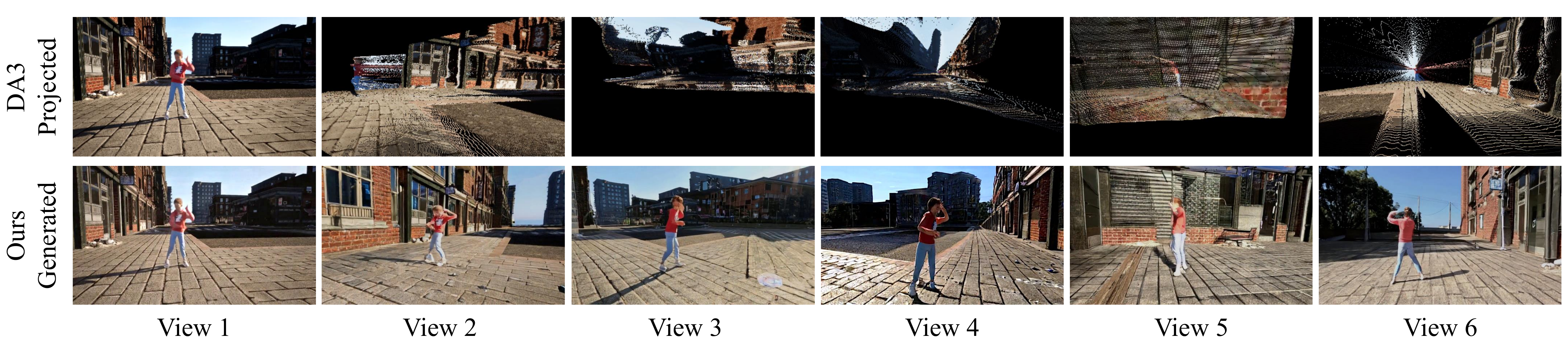}
    \vspace{-4mm}
    \caption{Comparison and illustration of depth-estimation method DA3~\cite{da3} and ours generated results. Directly using the depth-estimation method  leads to bad results over wide range of angles, but our model learns to refer its projection as condition to generate complete frames.}
    \vspace{-2mm}
    \label{fig:depth}
\end{figure}

\textbf{Spatial Guidance via Point Cloud Projection.}
To provide spatial guidance for target viewpoints, we first lift the input video into a dynamic point cloud via monocular depth estimation. For each frame \(t\), we estimate a depth map \(D_t\) and back-project the RGB pixels to 3D using the input camera pose, yielding a point cloud \(P_t\). Given a target camera \(\mathrm{cam}_v\), we render this point cloud to obtain a coarse image \(\tilde{I}_v\)  via perspective projection: \(\tilde{I}_v = \mathcal{R}(P_t, \mathrm{cam}_v)\), where \(\mathcal{R}\) denotes the rendering operation. This rendered image provides approximate geometry for the target view, which serves as a 4D-spatial prior for generation, see in Fig.\ref{fig:depth}.


We employ the 3D variational autoencoder with encoder $\mathcal{E}$ to project videos into a latent space.  The input video latent is denoted as $z_s = \mathcal{E}(V_s) \in \mathbb{R}^{f \times c \times h \times w}$, where $f$ stands for the number of frames, $c$ for the latent channels, and $h \times w$ for the spatial latent size. 
Similarly, $\tilde{I}_v$ is mapped to the  latent
    $z_v^{\mathrm{proj}} \in \mathbb{R}^{f \times c \times h \times w}$
using the same VAE.

To inject the source information into the generation process, we adopt a frame-dimension conditioning strategy. For each target viewpoint $v$, we first patchify its conditioning latent $z_v^{\mathrm{proj}}$ and its learnable target latent $z_v^{\mathrm{tgt}}$ (which is progressively denoised during inference) separately, then concatenate them along the frame dimension:

\begin{equation}
\begin{cases}
x_v^{\mathrm{tgt}} = \mathrm{patchify}(z_v^{\mathrm{tgt}}), \quad
x_v^{\mathrm{proj}} = \mathrm{patchify}(z_v^{\mathrm{proj}}), 
\\[1mm]
x_v = [x_v^{\mathrm{tgt}}, x_v^{\mathrm{proj}}]_{\mathrm{frame-dim}} \in \mathbb{R}^{2f \times s \times d},
\end{cases}
\end{equation}

where $z_v^{\mathrm{tgt}}$ is initialized as Gaussian noise during inference and gradually denoised to produce the final target video latent, $s = h \times w$ is the number of spatial tokens per frame, and $d$ is the token dimension. This concatenation effectively doubles the number of frames in the token sequence, allowing the model to jointly process the conditioning and target information.

For multiple target viewpoints, we stack their token sequences along the batch dimension to form a unified input:
\begin{equation}
X = [x_{v_1}, x_{v_2}, \dots, x_{v_{N_v}}]_{\mathrm{batch-dim}} \in \mathbb{R}^{N_v \times 2f \times s \times d}.
\end{equation}
$N_v$ is treated as the effective batch dimension, allowing all target viewpoints to be processed in parallel through the same DiT blocks with implicit cross-view aggregation via the shared projection conditioning. Furthermore, we'll introduce a fused  attention scheme between the view dimension $N_v$ and the temporal dimension $f$ to further enhance cross-view consistency in later sections.

\begin{figure}[t] 
    \centering
    \includegraphics[width=1\linewidth]{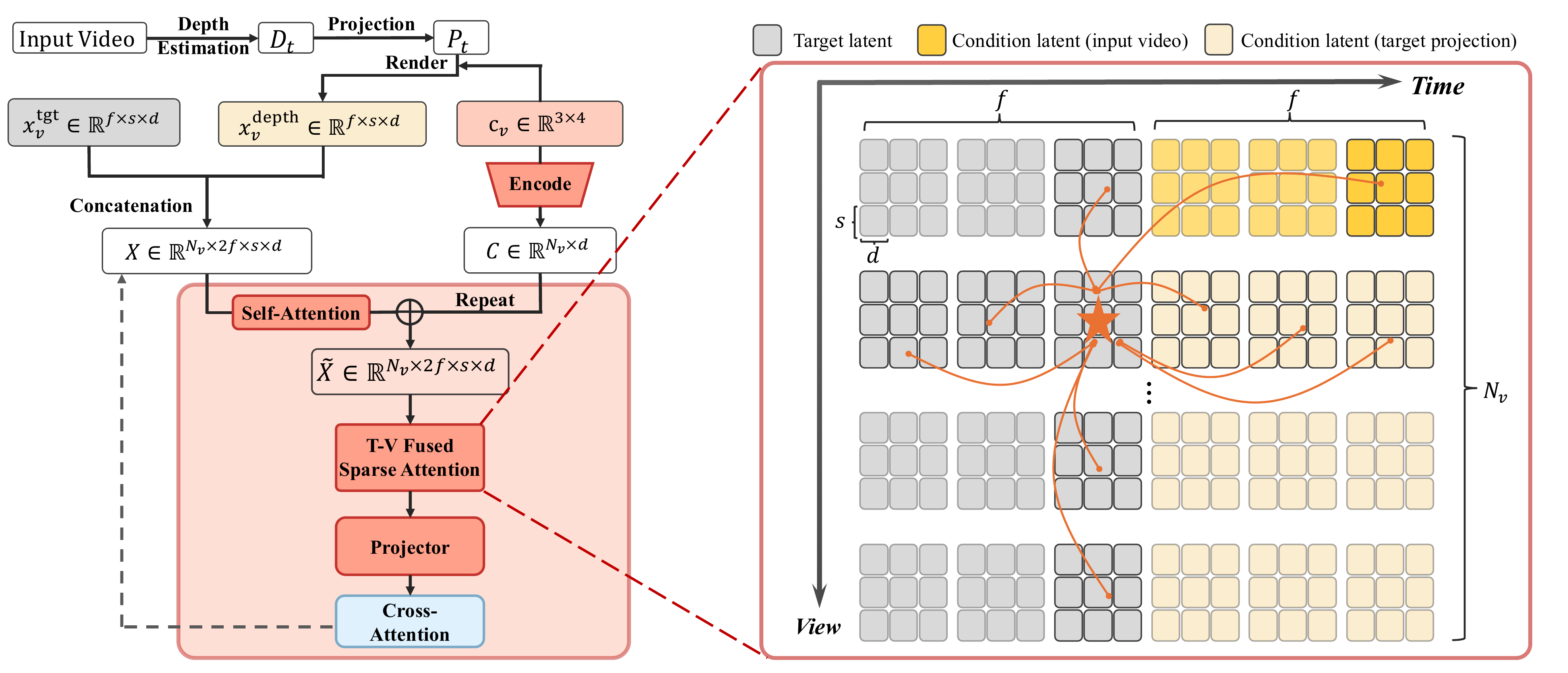}
    \caption{\textbf{Architectural details of the proposed video model.} \textit{Left}: projected condition and camera injection. The rendered latent and target latent are concatenated and processed by DiT blocks with camera embeddings \(C\) added after self-attention (Eq.\ref{add_cam}). \textit{Right}: visualization of our T-V fused sparse attention (Eq.\ref{eq:mask}). The mask strategy enables intra-view, intra-time and cross-half propagation.}
    \vspace{-2mm}
    \label{fig:architecture}
\end{figure}
\textbf{Explicit Camera Conditioning.} As shown in Fig.\ref{fig:architecture}, for each target viewpoint $v$, its camera parameters $\mathrm{c}_v = [R, t] \in \mathbb{R}^{3 \times 4}$ are defined relative to the reference viewpoint ($v=1$) consisting of rotation matrix $R$ and translation vector $t$. We encode each $\mathrm{c}_v$ via a learnable camera encoder $\mathcal{E}_c$ that maps $\mathrm{c}_v$ to the token dimension $d$, and stack them along the batch dimension to align with our token organization:
\begin{equation}
C = [\mathcal{E}_c(\mathrm{c}_{v_1}), \mathcal{E}_c(\mathrm{c}_{v_2}), \dots, \mathcal{E}_c(\mathrm{c}_{v_{N_v}})]_{\mathrm{batch-dim}} \in \mathbb{R}^{N_v  \times d},
\end{equation}
where each camera embedding is broadcast across all frames and spatial positions of its corresponding viewpoint. The camera-conditioned token sequence is then obtained by element-wise add to features after self-attention:

\begin{equation}\label{add_cam}
\tilde{X} = \operatorname{self\_attn}(X) + C \in \mathbb{R}^{N_v \times 2f \times s \times d}
,
\end{equation}
This injects consistent geometric information into all tokens of each target video before the  attention blocks. The camera encoder $\mathcal{E}_c$ is inserted into every DiT block and fine-tuned jointly with the model, enabling fine-grained camera control at multiple levels of the network. By conditioning each token on its corresponding viewpoint's camera parameters, the model learns to associate visual content with specific viewing geometries, facilitating the generation of multi-view consistent videos.  


\begin{figure}[t]
    \centering
    \includegraphics[width=1\linewidth]{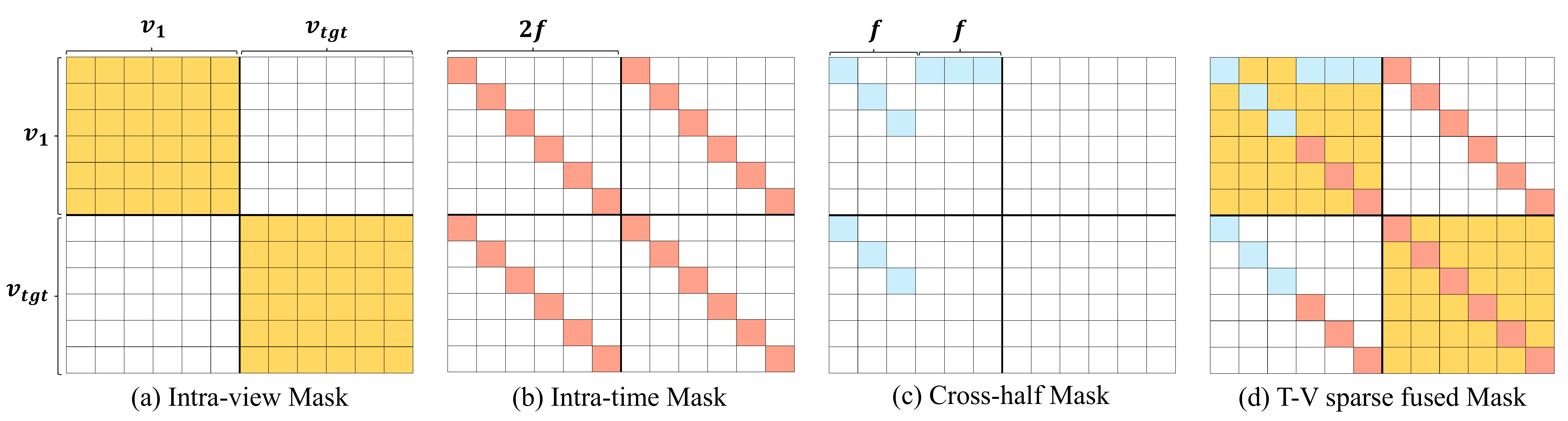}
    \caption{Illustration of masking strategy. (a) Intra-view mask: tokens within the same viewpoint attend across time, ensuring temporal coherence along each video, which is the traditional mode in video models.
(b) Intra-time mask: tokens from different viewpoints at the same timestamp attend to each other, enforcing cross-view geometric and appearance consistency.
(c) Cross-half mask: target frames attend to corresponding source frames at the same temporal offset, inheriting appearance and motion cues from the input video.
(d) T-V fused sparse mask: the combination of all.}
    \label{fig:mask}
\end{figure}
\textbf{Cross-View Synchronization via T-V Fused Sparse Attention.\label{fuseattn}}
The above modules enable per-view generation with target camera.  However, independently synthesizing each view breaks the cross-view geometric alignment and temporal synchronization required for 4D . Thus, we introduce a  fused sparse attention mechanism that enables direct information flow across both viewpoints and time during joint generation at once—ensuring a geometrically consistent and temporally coherent multi-view grid without additional parameters.

As illustrated in Fig.~\ref{fig:architecture}, our token representation $\tilde{X} \in \mathbb{R}^{N_v \times 2f \times s \times d}$ explicitly captures the 4D structure of the data. Each token is indexed by its viewpoint $v \in \{1,\dots,N_v\}$, temporal position $t \in \{1,\dots,2f\}$ (covering both target  and projected condition frames), and spatial coordinates $(p,q)$ with $p \in \{1,\dots,s\}$, $q \in \{1,\dots,d\}$, denoted as $\tilde{X}_{v,t,p,q}$. 

Within each DiT block, we compute query, key, and value matrices $\mathbf{Q}, \mathbf{K}, \mathbf{V} \in \mathbb{R}^{N \times d}$ from all tokens, where $N = N_v \cdot 2f \cdot s$ is the total number of tokens. We then apply a masked self-attention operation that restricts attention to tokens sharing either the same viewpoint or the same timestamp, or with input video:

\begin{equation}
\text{Attn}(\mathbf{Q},\mathbf{K},\mathbf{V}) = \text{SoftMax}\left(\mathbf{M} \odot \frac{\mathbf{Q}\mathbf{K}^\top}{\sqrt{d}}\right)\mathbf{V},
\end{equation}

where $\odot$ denotes element-wise multiplication and $\mathbf{M} \in \mathbb{R}^{N \times N}$ is a binary mask defined as:

\begin{equation} \label{eq:mask}
\mathbf{M}_{ij} = 
\begin{cases}
1, & \text{if } v_i = v_j \text{ or } t_i = t_j, \\
1, & \text{if } t_i < f \text{ and } t_j = t_i + f \text{ and } v_j=1, \\
0, & \text{otherwise},
\end{cases}
\end{equation}

\noindent with $i,j$ indexing tokens by their flattened 4D coordinates via $\text{Idx}(v,t,p,q)$. Here $v_i, t_i$ are recovered from index $i$ via the inverse mapping $\text{Idx}^{-1}(i) = (v_i, t_i, p_i, q_i)$, enabling the mask conditions to be directly evaluated using the original viewpoint and timestamp when computing attention scores $Q_i K_j^\top$. 

By indexing the same sequence along different dimensions, this masking strategy enables three critical types of information propagation, shown in Fig\ref{fig:mask}: (a)Intra-view ($v_i = v_j$) allows tokens within the same viewpoint to attend across time. (b)Intra-time propagation ($t_i = t_j$) enables tokens from different viewpoints at the same timestamp to interact directly. (c)Cross-half propagation ($t_i < f$ and $t_j = t_i + f$ and $v_j = 1$) establishes a direct pathway from the ground-truth input video ($v=1$) to the generated frames (since the first half stand for target and the later  for condition).

To incorporate the additional view dimension while remaining compatible with the pretrained 3D positional embeddings for $(t,p,q)$, we map the 4D coordinates to 3D by collapsing view and time dimensions:  $(v,t,p,q) \to (v \cdot T_{\text{max}} + t, p, q)$, where $T_{\text{max}}$ is the maximum temporal length. This treats the multi-view video as a single long sequence, allowing the original 3D positional encoding to be applied directly.


The attention mask has a density of $\frac{2f + N_v - 1}{2f \cdot N_v}$, which decreases as the number of views or frames increases, enabling efficient scaling. The sparse FlexAttention~~\cite{dong2024flexattentionprogrammingmodel} mechanism can be applied directly without introducing additional trainable parameters. In practice, we preserve the original self-attention and create a copy on which this mechanism operates, enabling cross-view interaction while maintaining the model’s original capabilities.

Finally, the aggregated features are then projected back to the spatial feature domain with a linear layer and residual connections.

\subsection{4D Reconstruction from Generated Multi-View Videos}
\label{sec:4dgs}

Given a set of generated multi-view videos $\{V_v\}_{v=1}^{N_v}=\{I_{v,t}\}$ with $N_v$ viewpoints and $f$ timestamps, we aim to recover a dynamic 4D scene representation that enables free-viewpoint rendering at arbitrary timestamps. We adopt 4D Gaussian Splatting (4D-GS) as our scene representation due to its explicit nature and real-time rendering capability.

\textbf{4D Gaussian Splatting.} 4D-GS~\cite{4dgs} extends static 3D Gaussian Splatting~\cite{3dgs} to dynamic scenes by maintaining a set of canonical 3D Gaussians $\mathcal{G} = \{g_i\}_{i=1}^N$, where each Gaussian is parameterized by position $\mu_i \in \mathbb{R}^3$, rotation $q_i \in \mathbb{R}^4$, scaling $s_i \in \mathbb{R}^3$, opacity $\alpha_i \in [0,1]$, and color $c_i \in \mathbb{R}^k$. A deformation field network $\mathcal{F}_{\mathrm{deform}}$ predicts per-frame offsets:

\begin{equation}
(\Delta \mu_i^t, \Delta q_i^t, \Delta s_i^t) = \mathcal{F}_{\mathrm{deform}}(\mu_i, t),
\end{equation}

where $t \in \{1,\dots,f\}$. Deformed Gaussians $\mu_i^t = \mu_i + \Delta \mu_i^t$, $q_i^t = q_i + \Delta q_i^t$, $s_i^t = s_i + \Delta s_i^t$ are rendered via differentiable splatting to produce images $\hat{I}_{v,t}$ at viewpoint $\mathrm{c}_v$.

\textbf{Flow Matching Distillation (FMD) Loss.}  
To leverage the generative prior of our pre-trained video diffusion model for regularizing the 4D representation, we adopt a flow-based distillation loss inspired by score distillation sampling (SDS)~~\cite{poole2022dreamfusion,4dgs}. The prior model, trained with the conditional flow matching objective, predicts a velocity field $v_\Theta(z_\tau, \tau, \mathrm{cond})$ for a noisy latent $z_\tau$ at time $\tau \in [0,1]$, given conditioning $\mathrm{cond}$ (including source video latent $z_s$, spatial latents $z_v^{\mathrm{proj}}$, and camera $\mathrm{c}_v$). A clean latent estimate is obtained as $f_\Theta(z_\tau, \tau, \mathrm{cond}) = z_\tau - \tau \, v_\Theta(z_\tau, \tau, \mathrm{cond})$.

During 4D Gaussian optimization, we sample a target viewpoint $v$ and render frames $\hat{I}_{v,t}$, which are encoded into latents $z_{v,t} = \mathcal{E}(\hat{I}_{v,t})$. These latents are corrupted with Gaussian noise according to the forward diffusion process:
\begin{equation}
z_{v,t}^\tau = (1-\tau) z_{v,t} + \tau \epsilon, \quad \epsilon \sim \mathcal{N}(0,I), \quad \tau \sim \mathcal{U}(0,1).
\end{equation}
The frozen prior model predicts the velocity $v_\Theta(z_{v,t}^\tau, \tau, \mathrm{cond})$, which is converted to a clean estimate $f_\Theta(z_{v,t}^\tau, \tau, \mathrm{cond})$. The FMD loss then compares this estimate with the original latent:
\begin{equation}\label{equation:fmd}
\mathcal{L}_{\mathrm{FMD}} = \mathbb{E}_{v,t,\tau,\epsilon} \left[ \omega(\tau) \left\| f_\Theta(z_{v,t}^\tau, \tau, \mathrm{cond}) - z_{v,t} \right\|_2^2 \right],
\end{equation}
where $\omega(\tau)$ is a time-dependent weighting function. This loss encourages rendered novel views to align with the prior distribution, filling occluded regions and enforcing consistency.

\textbf{Optimization.} Before reconstruction, we initialize the canonical 3D Gaussians $\mathcal{G}$ by fusing point clouds obtained from depth projections of the generated multi-view videos, and simultaneously perform re-cam-calibration using the same depth and pose estimates from DepthAnything-V3~\cite{da3}, ensuring alignment between the videos and the reconstructed scene.  The full objective for 4D reconstruction combines an image reconstruction loss, the proposed flow matching distillation loss, and deformation regularization terms:
\begin{equation}
\mathcal{L} = \mathcal{L}_{\mathrm{recon}} + \lambda_{\mathrm{FMD}} \mathcal{L}_{\mathrm{FMD}} + \lambda_{\mathrm{arap}} \mathcal{L}_{\mathrm{arap}} + \lambda_{\mathrm{rot}} \mathcal{L}_{\mathrm{rot}}.
\end{equation}
The reconstruction loss ensures fidelity to the generated observations following standard 4D-GS practices:
\begin{equation}
\mathcal{L}_{\mathrm{recon}} = \sum_{v,t} \left( \| \hat{I}_{v,t} - I_{v,t} \|_1 + \lambda_{\mathrm{ssim}} \mathcal{L}_{\mathrm{ssim}}(\hat{I}_{v,t}, I_{v,t}) \right),
\end{equation}
where $\hat{I}_{v,t}$ is the rendered image from 4D-GS and $I_{v,t}$ is the corresponding generated frame. The distillation loss $\mathcal{L}_{\mathrm{FMD}}$ (defined in Eq.\ref{equation:fmd}) leverages our pre-trained video diffusion prior to regularize novel views and smoothness. 

\section{Experiments}

\subsection{Dataset}
A major challenge in 4D generation research is the lack of established benchmarks and large-scale real-world datasets with wide viewpoint coverage. Existing datasets are often limited to synthetic renders or captures with narrow camera baselines, providing only partial observations that are insufficient for training and evaluating true surround-view generation models. To address this gap, we introduce \textbf{Real-MV-4D}, a large-scale dataset of synchronized multi-view videos captured in diverse real-world environments. Each scene is recorded by 6 widely spaced cameras surrounding the entire scene, providing full-scope observations essential for learning  4D scene generation.

The dataset comprises over 10,000 hours of footage, with 7,000+ hours of single-person actions, 2,200+ hours of two-person interactions, and 800+ hours of multi-person complex activities. All videos are captured at 1080p resolution with continuous actions exceeding 60 minutes per clip. It spans 2,000+ indoor and outdoor scenes featuring 6,000+ subjects, yielding over 500,000 distinct action instances.

For our experiments, we use over 60,000 video clips sampled from 3,000 scenes for training, each clip is an 81-frame subsequence cropped from the original scene videos, where we select the 20 most dynamic segments based on motion magnitude to ensure sufficient temporal variation. To establish a benchmark for evaluation, we select 500 diverse scenes covering varied environments, subject counts, and action types as our test set. This allows systematic assessment of model performance on generating truly full-scope  4D scenes across different scenarios.

\subsection{Settings}
\noindent \textbf{Training Strategy.} 
For every clip from our dataset, we randomly select one video as the source viewpoint $v_1$ and treat the remaining five, in random order, as target viewpoints ${v}_{t=2}^{6}$ to be generated. This randomization exposes the model to diverse relative camera transformations, complementing the inherent diversity of our unaligned multi-camera setup. Each batch contains all 6 videos from a scene, enabling joint learning of multi-view consistency.

At inference, the model can generate any number of target views, with the source video retained as the first row of desired 4D video grid to provide appearance and motion cues. Although the model is trained with six views jointly, inference supports an arbitrary number of videos. To fully leverage the T–V fused attention mechanism, we recommend generating multiple views together based on available computational resources.

We build our models upon Wan2.2-TI2V-5B~~\cite{wan} and employ DepthAnything-V3 (DA3)~~\cite{da3} to provide the projection conditioning. Training is performed for 10k steps at a resolution of 384×672 using 8 H100 GPUs. The camera encoder and projector are initialized to zero.

\noindent \textbf{Baselines.}  We compare our method against three representative baselines covering both camera-controlled video-to-video generation and 4D scene generation paradigms. ReCamMaster~\cite{recammaster} is a camera-controlled video re-rendering framework that reproduces dynamic scenes at novel camera trajectories. Free4D~\cite{liu2025free4d} is a tuning-free approach that generates 4D scenes from a single image by animating the input and refining multi-view videos through point-guided denoising. DimensionX~\cite{dimensionx} generates 3D/4D scenes by decoupling spatial and temporal factors via specific LoRAs.  For a fair comparison, we provide the same fixed-viewpoint videos as input and ReCamMaster’s camera trajectories are modified to match our fixed viewpoints. For methods with available reconstruction code, we compare using their reconstructed outputs. For others, or when reconstruction fails due to low-quality video generation,  we use the generated videos  for  evaluation. This is reasonable since poor-quality  videos across large viewpoint ranges  definitely yield poor full-scope 4D reconstructions.


\noindent \textbf{Metrics.}  In order to comprehensively evaluate the performance of the proposed method, we divide the evaluation metrics into two categories: Visual Quality and View Synchronization.
\textbf{Visual Quality} assesses the intrinsic quality of the rendered videos. We compute FID and FVD to measure the distributional similarity between generated and real data. In addition, we calculate CLIP-F, defined as the CLIP similarity between adjacent frames of generated videos, to evaluate temporal coherence. 
\textbf{View Synchronization} measures the consistency between input and output videos. We compute Mat.Pix. using GiM~\cite{gim}, which counts the number of matched feature points between input and output frames at the same timestamp. We also compute FVD-V between the input and output videos, and CLIP-V between corresponding frames.
Furthermore, we evaluate the rendered videos on VBench to obtain a comprehensive benchmark score.

\begin{figure}[!]
    \centering
    \includegraphics[width=1\linewidth]{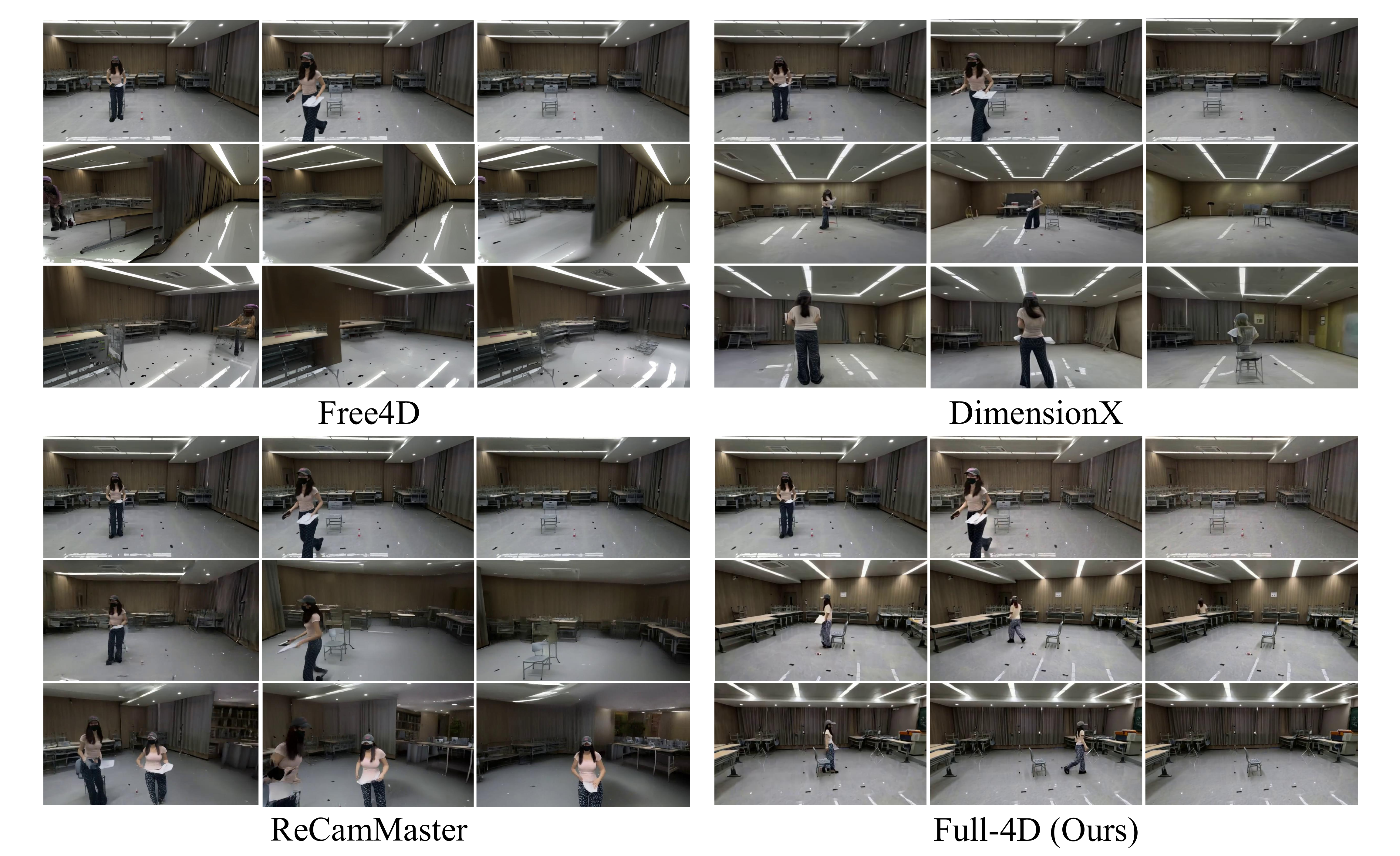}
    
    \caption{\textbf{Indoor scene comparison.}
Across front and two roughly 90° side views, baselines struggle with spatial distortions and inconsistent human appearances, while our method preserves sharp geometry and coherent full-scope
 structure.}
\label{fig:indoor}
\includegraphics[width=1\linewidth]{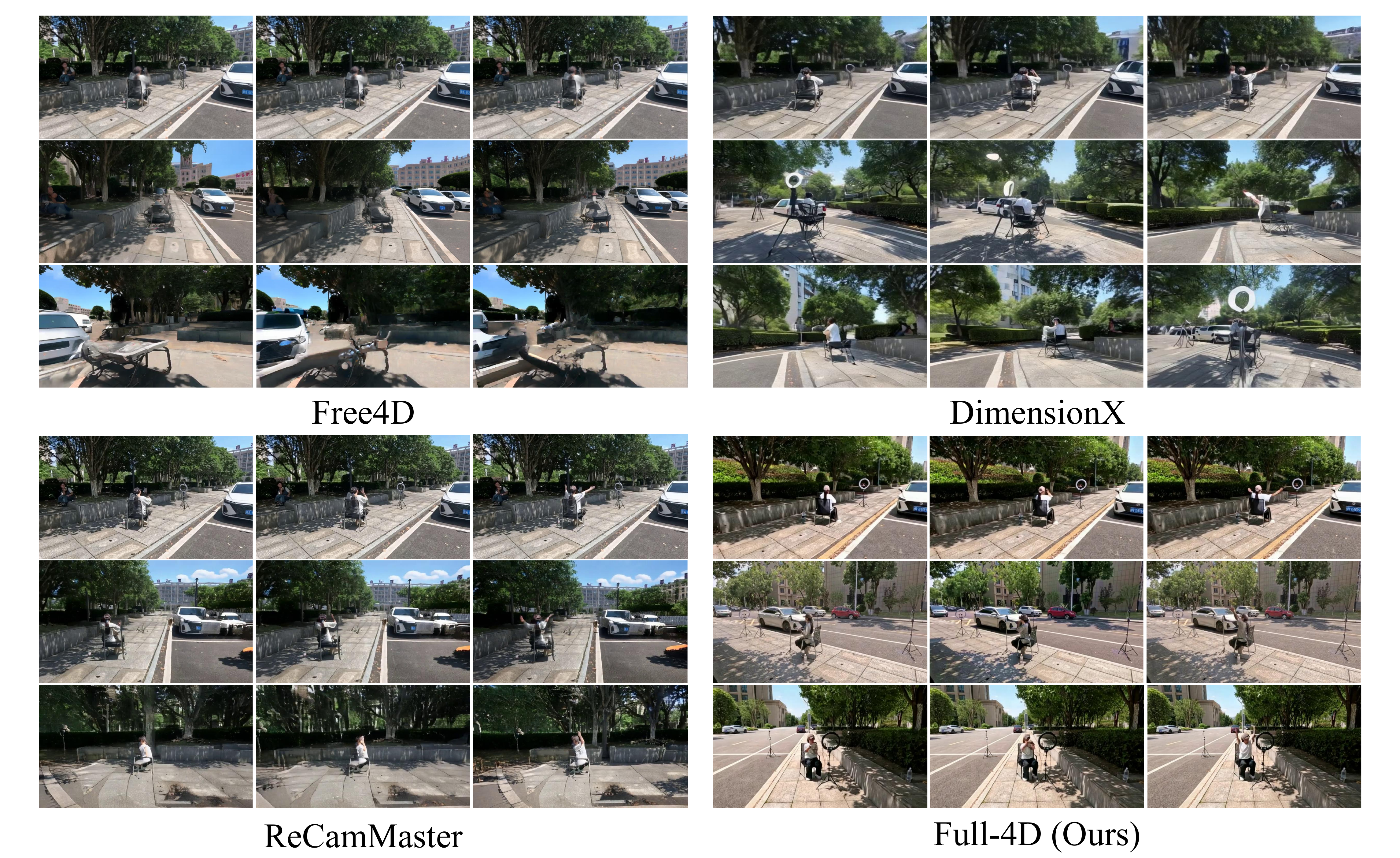}
    \caption{\textbf{Outdoor scene comparison.} Across large front-to-back viewpoint changes, baselines exhibit blur, spatial distortions, and temporal inconsistencies, while our method preserves realistic appearance, accurate geometry, and coherent structure.}
    \label{fig:outdoor}
\end{figure}

\begin{table}
\centering
\caption{Quantitative comparison on visual quality and view consistency.}
\label{tab:quality}
\resizebox{\textwidth}{!}{%
\begin{tabular}{lccc|ccc|c}
\toprule
& \multicolumn{3}{c|}{Visual Quality} & \multicolumn{3}{c}{View Synchronization} & \\
\cmidrule(lr){2-4} \cmidrule(lr){5-7}
\multirow{-2}{*}{Method} & FID $\downarrow$ & FVD $\downarrow$ & CLIP-F $\uparrow$ & Mat. Pix.(K) $\uparrow$ & FVD-V $\downarrow$ & CLIP-V $\uparrow$ & \multirow{-2}{*}{View Consistency$\uparrow$} \\
\midrule
Free4D~\cite{liu2025free4d} & 115.74 & 741.88 & 96.74 & 109.01 & 680.65 & 90.63 & 0.674\\
DimensionX~\cite{dimensionx} & 98.69 & 577.84 & 92.72 & 365.53 & 560.87 & 91.37 & 0.500\\
ReCamMaster~\cite{recammaster} & 62.38 & 360.86 & 98.36 & 448.72 & 383.28 & 90.50 & \textbf{0.736}\\
Full-4D(Ours) & \textbf{35.12} & \textbf{175.29} & \textbf{99.74} & \textbf{540.92} & \textbf{140.60} & \textbf{93.17} & 0.719\\
\bottomrule
\end{tabular}%
}
\end{table}

\begin{table}
\centering
\caption{Quantitative VBench evaluation of rendered 4D videos.}
\label{tab:vbench}
\footnotesize 
\begin{tabular}{lcccccc}
\toprule
Method & Aesthetic & Background & Imaging & Motion & Subject & Temporal \\
& $\uparrow$ & Consistency $\uparrow$ & Quality $\uparrow$ & Smoothness $\uparrow$ & Consistency $\uparrow$ & Flickering $\uparrow$ \\
\midrule
Free4D~~\cite{liu2025free4d} & 41.34 & 90.21 & 61.43 & 96.36 & 89.00 & 95.83 \\
DimensionX~~\cite{dimensionx} & 46.07 & 89.79 & 64.35 & 91.12 & 82.70 & 89.50 \\
ReCamMaster~~\cite{recammaster} & \textbf{48.99} & 88.44 & 65.15 & 99.13 & 87.11 & 97.92 \\
\midrule
Full-4D(Ours) & 46.26 & \textbf{96.13} & \textbf{70.80} & \textbf{99.64} & \textbf{97.77} & \textbf{99.63} \\
\bottomrule
\end{tabular}
\end{table}

\begin{table}
\centering
\footnotesize 
\caption{Quantitative comparison on Real-MV-4D and DyCheck datasets.}
\label{tab:psnr}
\begin{tabular}{lccc|ccc}
\toprule
& \multicolumn{3}{c|}{Real-MV-4D} & \multicolumn{3}{c}{DyCheck~\cite{dycheck}} \\
\cmidrule(lr){2-4} \cmidrule(lr){5-7}
\multirow{-2}{*}{Method} & PSNR $\uparrow$ & SSIM $\uparrow$ & LPIPS $\downarrow$ & PSNR $\uparrow$ & SSIM $\uparrow$ & LPIPS $\downarrow$ \\
\midrule
Free4D~\cite{liu2025free4d} & 13.16 & 0.436 & 0.656& 11.83 & 0.313 & 0.652 \\
DimensionX~\cite{dimensionx} & 13.32 & 0.372 & 0.584& 11.59& 0.338& 0.672\\
ReCamMaster~\cite{recammaster} & 13.45 & 0.396 & 0.613& 12.44 & 0.376 & 0.546 \\
TrajectoryCrafter~\cite{trajectorycrafter} & 13.60& 0.432& 0.592& 14.34 & 0.450 & 0.495 \\
Full-4D(Ours) & \textbf{15.50} & \textbf{0.486} & \textbf{0.654} & \textbf{14.59}& \textbf{0.465}& \textbf{0.433}\\
\bottomrule
\end{tabular}
\end{table}

\subsection{Comparisons with baselines}
\noindent \textbf{Qualitative Results.} Fig.~\ref{fig:indoor} and Fig.~\ref{fig:outdoor} show qualitative comparisons on indoor and outdoor scenes respectively. The indoor set includes the front view and two roughly 90° side views in Fig.~\ref{fig:indoor}: Free4D~~\cite{liu2025free4d} exhibits poor spatial continuity with noticeable scene distortions especially behind walls, DimensionX~~\cite{dimensionx} produces multi-faced human instances where the same person appears frontal from different viewpoints, and ReCamMaster~~\cite{recammaster} generates multiple duplicated human instances, reflecting weak multi-view consistency. All baselines show distortions throughout the scene geometry. The outdoor set in Fig.~\ref{fig:outdoor} covers large-angle front-to-back transformations, where Free4D~~\cite{liu2025free4d} is noticeably blurred and all baselines struggle with spatial distortions and inconsistent appearances under viewpoint changes. In contrast, our method maintains sharp geometry, photorealistic textures, and consistent multi-view structure across full-scope views, demonstrating robust performance in both indoor and outdoor scenes.  Additional results (Fig.~\ref{fig:teaser} and Fig.~\ref{fig:depth}) demonstrate that despite being trained on our dataset, our method generalizes well to unseen scenes beyond the training split.

\noindent \textbf{Quantitative Results.}  
Table~\ref{tab:quality} shows that our method consistently outperforms all baseline approaches across both Visual Quality and View Synchronization metrics. In terms of visual quality, our model achieves the best performance on FID and FVD, indicating that the generated videos are closer to the distribution of real multi-view videos. The higher CLIP-F score further demonstrates stronger temporal coherence, validating the effectiveness of the proposed fused attention mechanism in maintaining stable frame-to-frame semantics.
For view synchronization, our method also achieves the best results on Mat. Pix., FVD-V, and CLIP-V, indicating better geometric and semantic alignment between the rendered and input videos. WE also provide View Consistency score from 4DWorldBench~\cite{lu20254dworldbench,duan2025worldscore} calculated by reprojection error. These results show that our approach preserves cross-view correspondences more effectively, leading to improved multi-view consistency.
The VBench results in Table~\ref{tab:vbench} further confirm these advantages. While the aesthetic score is comparable to ReCamMaster~~\cite{recammaster}, our method achieves the best performance on all other metrics. In particular, the significantly higher background consistency and subject consistency demonstrate that our approach better preserves scene structure and object identity across viewpoints and time, which is crucial for full-scope 4D scene generation. We also  provide reference metrics for all baselines for comparison on both of 
our dataset and the iPhone Dycheck dataset (Block, Paper,
Spin, Teddy, excluding Apple due to the lack of a contigu- 
ous 81-frame overlap between its two fixed views) in Tab.\ref{tab:psnr}.

\section{Ablation}
\begin{figure}[!t]
    \centering
    \includegraphics[width=0.8\linewidth]{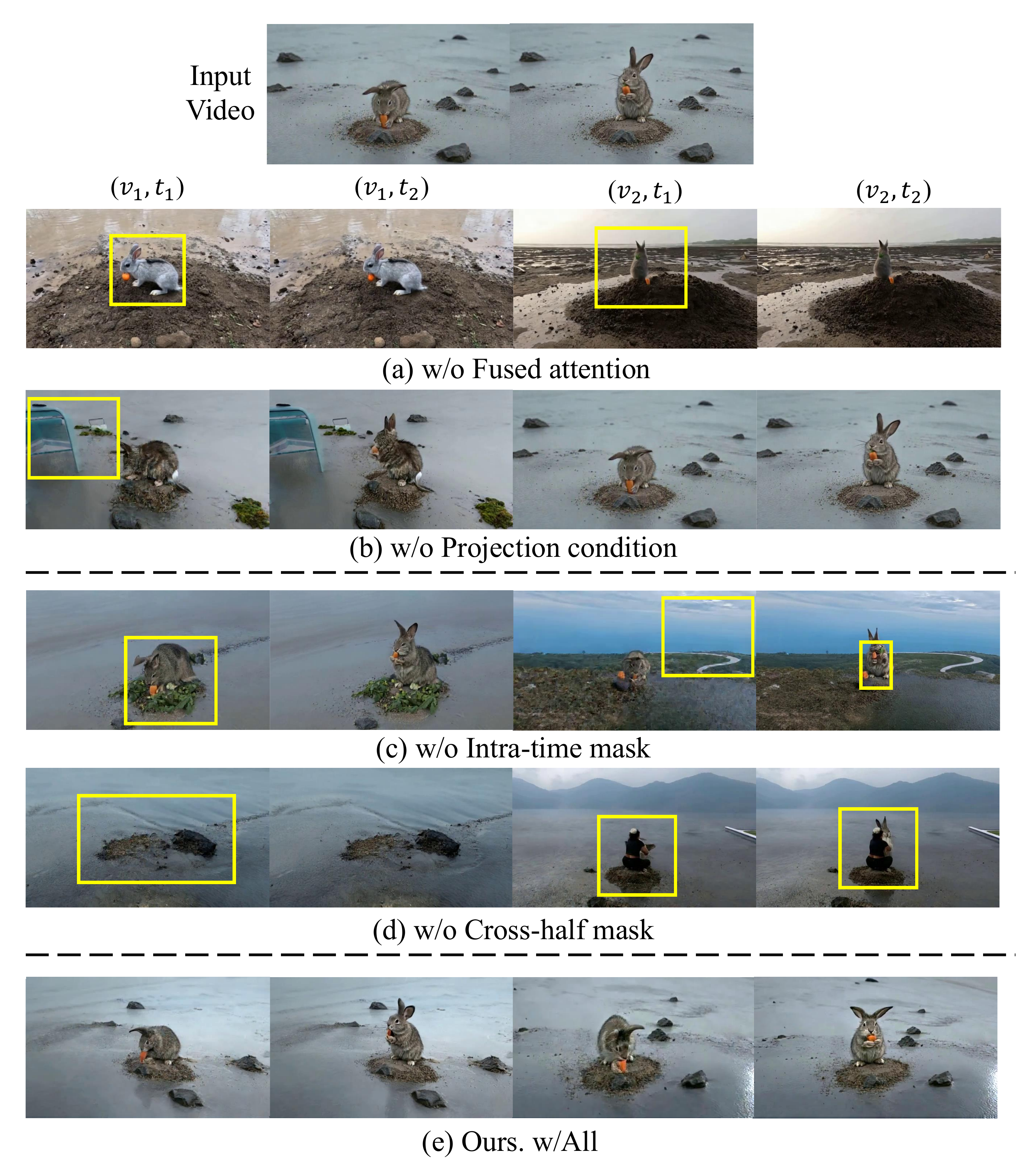}
    \caption{Ablations of key componets of our multi-view video generation model. Each row shows a variant with a specific component removed, displaying two frames from two different viewpoints. Yellow boxes highlight degradation caused by the ablation.}
    \label{fig:ablation}
\end{figure}
Below we provide detailed ablation studies on the key components of our framework, covering both  multi-view video generation  and  4D reconstruction.

\subsection{Ablation on Multi-view Video Generation}
We first analyze the design choices in the multi-view video diffusion model, including T–V fused attention and spatial-aware projection conditioning. All variants are trained under identical settings, and differences are solely due to the removed components.

We begin by removing the fused time–view sparse attention and reverting to the original per-video self-attention, where each target view is generated independently. In this case, the model lacks explicit cross-view interaction during denoising. As shown in Fig.\ref{fig:ablation}(a), the generated multi-view results deviate significantly from the input video. The synthesized views show poor preservation of subject identity and scene structure across viewpoints, with only background regions near the projection-conditioned areas remaining partially similar. This demonstrates that independently generating each view cannot recover a synchronized T×V grid, leading to severe cross-view inconsistency.

Next, we remove the spatial-aware projection conditioning while retaining fused attention. As illustrated in Fig.\ref{fig:ablation}(b), although the generated content still loosely follows the input video due to fused propagation, the background becomes unstable across views. Moreover, the model struggles to learn the intended camera transformations, often simply replicating the input viewpoint across all target views instead of synthesizing novel perspectives. The model lacks explicit spatial anchors for novel viewpoints, resulting in unreasonable layouts and failed view transformation. This confirms that projection conditioning plays a critical role in stabilizing large-viewpoint-change synthesis by providing coarse but effective geometric and viewpoint guidance.

We further ablate the fused attention by separately removing the intra-time mask and the cross-half mask (we retain the standard intra-view attention as it is a conventional component in video diffusion models). When the intra-time mask is removed (Fig.\ref{fig:ablation}(c)), subject motions become inconsistent across viewpoints at the same timestamp, and in some cases the background also loses cross-view coherence. When the cross-half mask is removed (Fig.\ref{fig:ablation}(d)), the generated views lose reference to the input video, leading to the most severe identity degradation: the input rabbit may disappear or even be replaced by a human-like figure. These results indicate that the intra-time mask enforces cross-view motion synchronization, while the cross-half mask is essential for preserving subject identity and semantic consistency from the source video. Notably, a naive full attention over all tokens across all views and timestamps would theoretically capture all these interactions, but it is computationally prohibitive in both memory and time. Our sparse masking strategy achieves the same critical information flow while enabling efficient training.

These qualitative observations are further supported by the quantitative results in Tab.~\ref{tab:ablation_sync}. 
Removing any key component leads to consistent degradation in cross-view synchronization metrics. 
In particular, discarding fused attention results in a substantial drop in Mat. Pix. and a large increase in FVD-V, indicating severe synchronization failure across views. 
Similarly, removing projection conditioning or either attention mask also degrades performance compared to the full model. 
Our complete model achieves the best performance across all synchronization metrics, demonstrating the effectiveness of each design choice.
\begin{figure}
    \centering
    \includegraphics[width=1\linewidth]{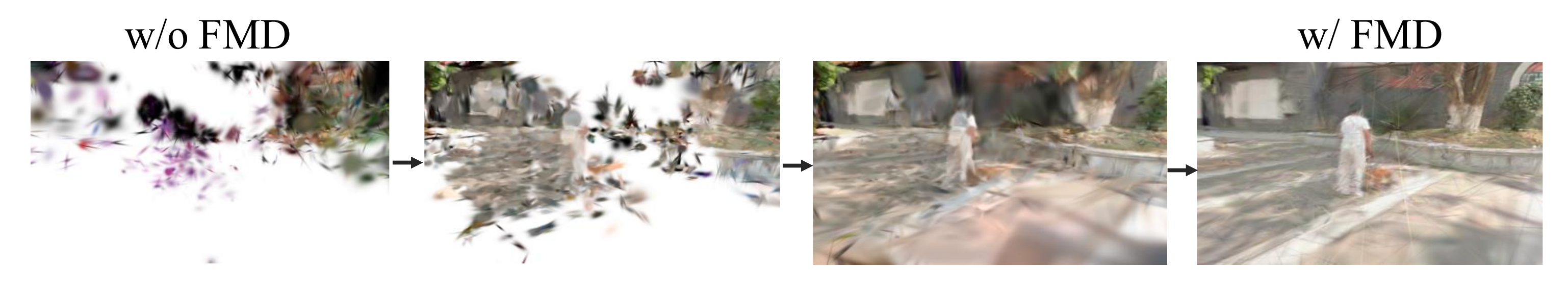}
    \caption{\textbf{FMD ablation during 4DGS optimization.} Renderings from a representative interpolated viewpoint at different training stages. Without FMD (left), the rendered results exhibit missing structures and sparse Gaussian-like artifacts due to insufficient supervision. With FMD (right), the scene progressively forms coherent and complete geometry, demonstrating the effectiveness of FMD in enforcing cross-view consistency.}
    \label{fig:fmd}
\end{figure}
\begin{table}[h]
\centering
\caption{Ablation study of key components on multi-view video generation.}
\label{tab:ablation_sync}
\footnotesize%
\begin{tabular}{lccc}
\toprule
Method & Mat. Pix.(K) $\uparrow$ & FVD-V $\downarrow$ & CLIP-V $\uparrow$ \\
\midrule
w/o Fused Attention & 101.35 & 528.47 & 90.84 \\
w/o Projection Conditioning & 468.72 & 252.63 & 92.26 \\
w/o Intra-time Mask & 389.54 & 210.72 & 91.02 \\
w/o Cross-half Mask & 165.41 & 318.65 & 89.47 \\
Full Model (Ours) & \textbf{540.92} & \textbf{140.60} & \textbf{93.17} \\
\bottomrule
\end{tabular}%
\end{table}

\begin{table}[!h]
\footnotesize 
\centering
\caption{Ablation study of FMD during Gaussians optimization.}
\label{tab:ablation_fmd}
\footnotesize%
\begin{tabular}{lccc}
\toprule
Method & Mat. Pix.(K) $\uparrow$ & FVD-V $\downarrow$ & CLIP-V $\uparrow$ \\
\midrule
w/o FMD & 256.28 & 768.57 & 83.84 \\
Full Model (Ours) & \textbf{509.73} & \textbf{187.69} & \textbf{90.93} \\
\bottomrule
\end{tabular}%
\end{table}
\subsection{Ablation on Flow Matching Distillation}
For the FMD ablation, we remove the FMD constraint during optimization and train the 4DGS representation using only the six RGB viewpoints. As shown in Fig.\ref{fig:fmd}, without FMD, the Gaussians receive little supervision at non-RGB poses, resulting in missing content or sparse Gaussian-like artifacts in the rendered images. With FMD enabled, the intermediate non-RGB viewpoints progressively form coherent and continuous renderings, demonstrating improved geometric completeness across interpolated views using the video prior. For quantitative evaluation, metrics are computed over both the original RGB viewpoints and the interpolated FMD viewpoints in Tab.\ref{tab:ablation_fmd}.
\section{Conclusion}

Generating dynamic 4D scenes from a single-view video remains challenging because coherent 4D modeling requires continuity across both time and viewpoints. In this work, we address this problem through three key components.
First, we introduce Real-MV-4D, a large-scale dataset of synchronized multi-view videos.
Second, we develop a multi-view video diffusion model that generates a synchronized $T\times V$ grid of videos from a single-view video, enabled by spatial-aware projection conditioning and a fused time–view sparse attention mechanism that allows information exchange across view,  time and input video.
Third, we lift the generated multi-view videos into an explicit 4D Gaussian representation, where a flow matching distillation loss leverages the pretrained video prior to regularize the optimization.
Together, these components enable coherent 4D scene generation from a single-view input.

\clearpage

\appendix

\begin{center}
    \vspace*{1em}
    {\Huge \bfseries Supplementary Material \par}
    \vspace{1em}
\end{center}

\section{Real-MV-4D Dataset}
Real-MV-4D is derived from a large-scale synchronized multi-camera capture project conducted by our industrial–academic research team. Each scene is captured by six hardware-synchronized RGB cameras positioned around the activity area. The setup provides near 360° coverage with substantial viewpoint variation. All video streams are strictly aligned at the frame level and captured at 1080p resolution, enabling reliable cross-view correspondence over long temporal durations. Camera timestamps are aligned to a central server via NTP, yielding sub-frame synchronization error ($<33$ms). We use a standard pinhole model, calibrate intrinsics and initial extrinsics with a chessboard, and globally refine extrinsics via dense cross-view pixel alignment to reduce multi-view noise.

The dataset comprises over 2,000 indoor and outdoor scenes with more than 6,000 subjects, totaling over 10,000 hours of synchronized multi-view video. All recordings are captured in real-world settings under natural lighting and complex scene conditions. They cover diverse environments and activities, ranging from single-person actions to multi-person and human–object interactions. This large-scale synchronized multi-view supervision supports the design and training of our 4D T$\times$V generation framework.

For our experiments, we sample 60,000 temporally aligned six-view clips from over 3,000 scenes for training and reserve 500 disjoint scenes for testing to evaluate generalization to unseen environments. We further refine the camera parameters using depth-based geometric alignment to improve cross-view projection consistency during 4D training.

Existing 4D-related datasets primarily rely on synthetic data, object-centric captures, or monocular videos with small camera motion. Such settings lack synchronized multi-view observations and limit the ability to model large viewpoint changes under realistic scene dynamics.
In contrast, Real-MV-4D consists of real-world recordings with hardware-synchronized multi-view capture over time. It provides jointly aligned spatial and temporal observations across diverse scenes and activities, enabling learning under both wide-baseline viewpoint variation and complex real-world dynamics.

\begin{figure}[h]
    \centering
\includegraphics[width=1\linewidth]{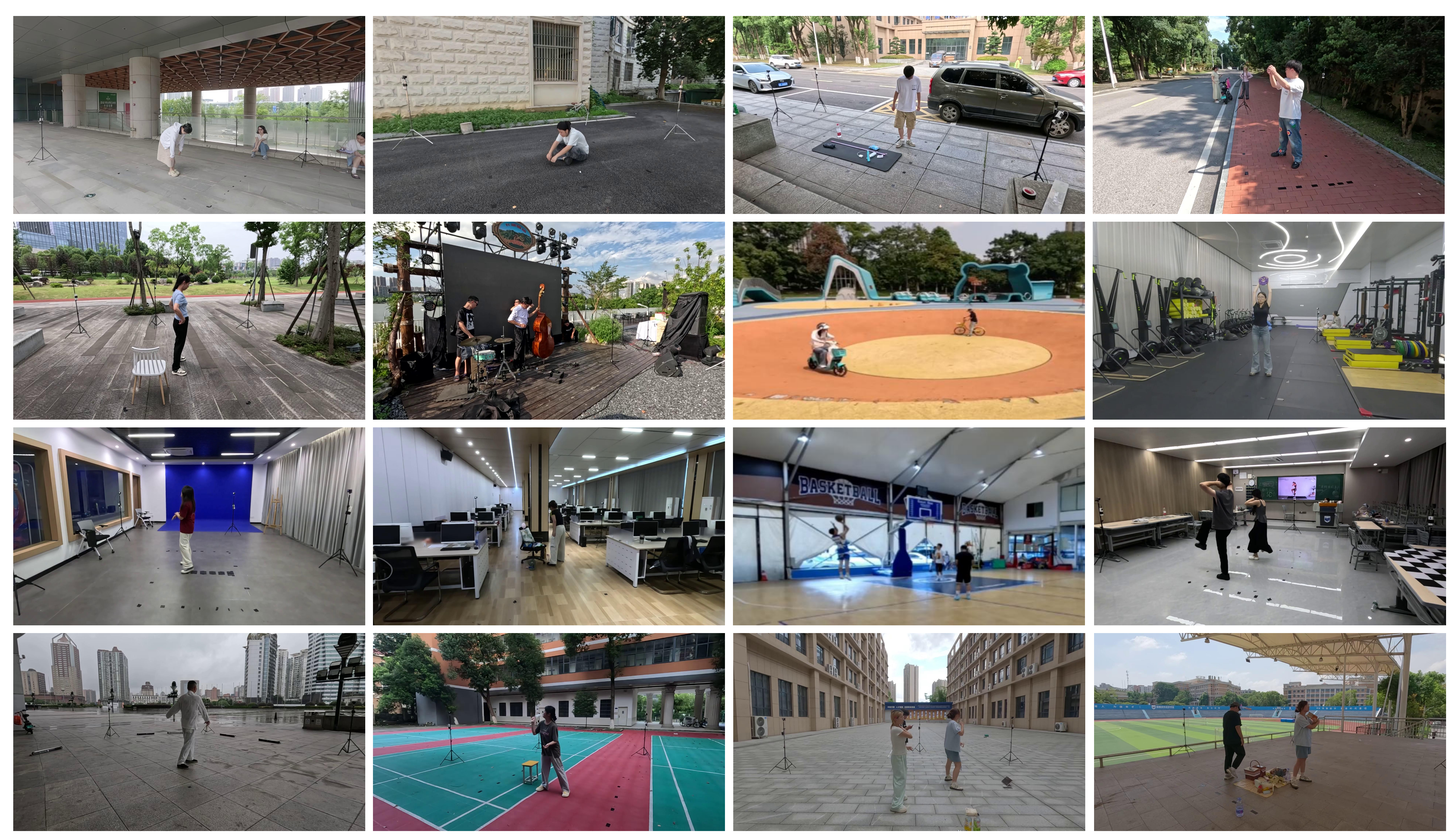}
    \caption{Samples from the Real-MV-4D, which covers a wide variety of indoor and outdoor scenes, diverse human activities, and rich interactions, demonstrating the large-scale diversity essential for training generalizable 4D  models.}
    \label{fig:dataset}
\end{figure}

Due to commercial and privacy considerations, we plan to partially release curated training or testing subsets after publication. We show some samples in Fig.\ref{fig:dataset}.
\section{Experiment Details}
\begin{figure}[t]
    \centering
    \includegraphics[width=0.75\linewidth]{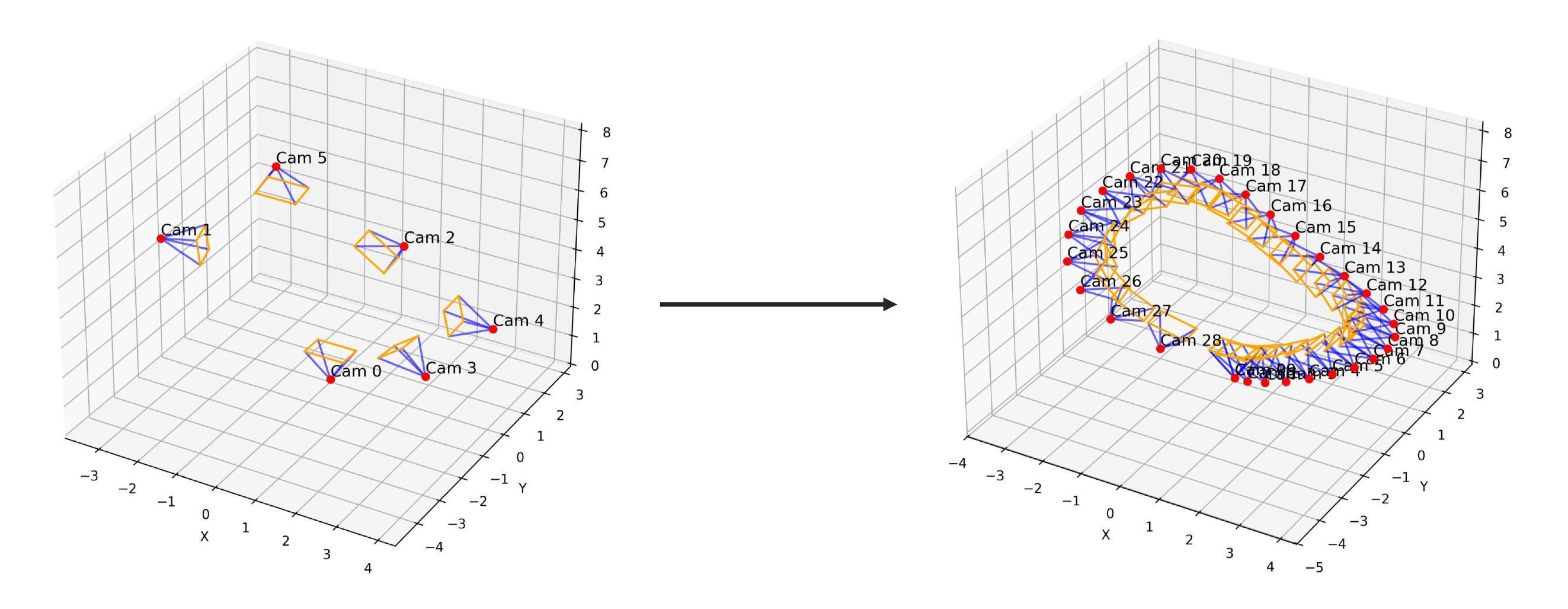}
    \caption{\textbf{Illustration of camera pose interpolation.} Six RGB viewpoints from the dataset are used to supervise 4D Gaussian reconstruction (left). Based on these views, twenty-eight additional camera poses are generated through interpolation to form a smooth trajectory around the scene center for FMD loss computation and visualization (right). For clarity of presentation, only 28 interpolated poses are shown here, while 120 interpolated poses are used in the actual experiments.}
    \label{fig:camera}
\end{figure}

\noindent\textbf{Interpolated Camera Poses} 
For each scene, we use a set of six RGB viewpoints from the  dataset. These views are used to generate RGB video sequences for supervising the 4D Gaussian reconstruction.
Additional camera poses for FMD loss computation are obtained by interpolating between these six viewpoints, as illustrated in Fig.\ref{fig:camera}. To ensure spatial continuity before interpolation, we first sort the camera poses $\{\mathbf{T}_i^{\text{w2c}}\}_{i=1}^{6}$ according to their azimuthal angles in the horizontal plane. Specifically, we compute the angle of each camera center $\mathbf{p}_i = (\mathbf{T}_i^{\text{c2w}})_{:3,3}$ relative to the scene center $\mathbf{c} = \frac{1}{6}\sum_{i=1}^{6}\mathbf{p}_i$ as:
\begin{equation}
\phi_i = \arctan2(z_i - c_z, x_i - c_x),
\end{equation}
where $(x_i, z_i)$ denote the horizontal coordinates. The cameras are reordered in ascending order of $\phi_i$ to form a spatially continuous sequence, which serves as the basis for subsequent pose interpolation.
For the sorted camera positions $\{\mathbf{p}_i\}_{i=1}^{6}$, we apply cubic spline interpolation to generate smooth trajectories with $C^2$ continuity. To ensure loop closure, we append $\mathbf{p}_1$ to the end, forming $\{\mathbf{p}_i\}_{i=1}^{7}$ with $\mathbf{p}_7 = \mathbf{p}_1$. The interpolated position at time $t \in [0, 6]$ is:
\begin{equation}
\mathbf{p}(t) = \text{CubicSpline}(\{0, 1, \ldots, 6\}, \{\mathbf{p}_1, \ldots, \mathbf{p}_7\})(t),
\end{equation}
here $t$ is uniformly sampled at 120 points: $t_j = 6(j-1)/119$ for $j = 1, \ldots, 120$.
Camera orientations are represented as quaternions $\{\mathbf{q}_i\}_{i=1}^{7}$ (with $\mathbf{q}_7 = \mathbf{q}_1$ for closure) and interpolated via Spherical Linear Interpolation (SLERP) to maintain smooth rotation transitions on the $SO(3)$ manifold:
\begin{equation}
\mathbf{q}(t) = \text{Slerp}(\{0, 1, \ldots, 6\}, \{\mathbf{q}_1, \ldots, \mathbf{q}_7\})(t).
\end{equation}
SLERP ensures the shortest geodesic path between rotations, defined as:
\begin{equation}
\text{Slerp}(\mathbf{q}_a, \mathbf{q}_b, \alpha) = \frac{\sin((1-\alpha)\theta)}{\sin\theta}\mathbf{q}_a + \frac{\sin(\alpha\theta)}{\sin\theta}\mathbf{q}_b,
\end{equation}
where $\theta = \arccos(\mathbf{q}_a \cdot \mathbf{q}_b)$ and $\alpha = t - \lfloor t \rfloor$.

\noindent\textbf{4D Reconstruction} Because the large-scale video generation model needs to compute FMD by rendering the entire video sequence and preserving the full computational graph, we use two NVIDIA H100 GPUs during the 4D reconstruction stage. One GPU computes the FMD loss with the frozen video generation model, while the other optimizes the dynamic Gaussian representation. 
Furthermore, due to GPU memory constraints, at each training iteration we randomly sample one interpolated camera pose from the generated camera poses and render the entire video sequence from that viewpoint for FMD loss computation.  With access to more GPU resources, simultaneous optimization across multiple novel viewpoints would be feasible, enabling the model to better leverage its capacity for multi-view consistency.
To further limit GPU memory usage, we constrain the current number of Gaussian points with an upper bound of 120,000 during densification and growth, and apply pruning when the number of Gaussians exceeds 80,000. The opacity threshold for pruning is set to 0.015.

\section{More results}
\begin{figure}
    \centering
    \includegraphics[width=1\linewidth]{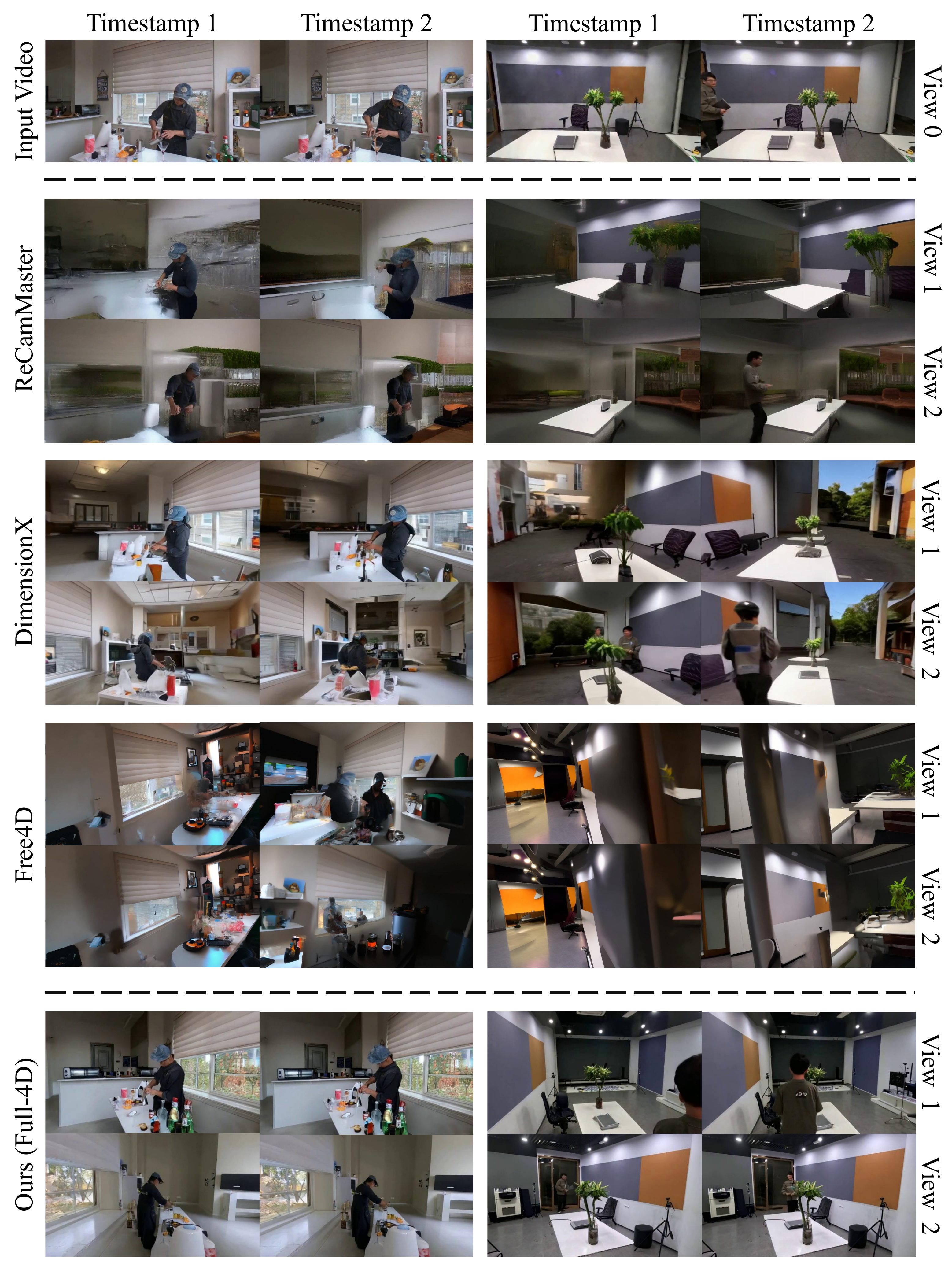}
    \caption{\textbf{Additional qualitative comparisons.} This figure presents two representative cases (left and right), where all methods are evaluated using the same input video. For each case, the first and second columns correspond to two different timestamps, and the first and second rows correspond to two significantly different viewpoints.}
    \label{fig:case1}
\end{figure}

\begin{figure}
    \centering
    \includegraphics[width=1\linewidth]{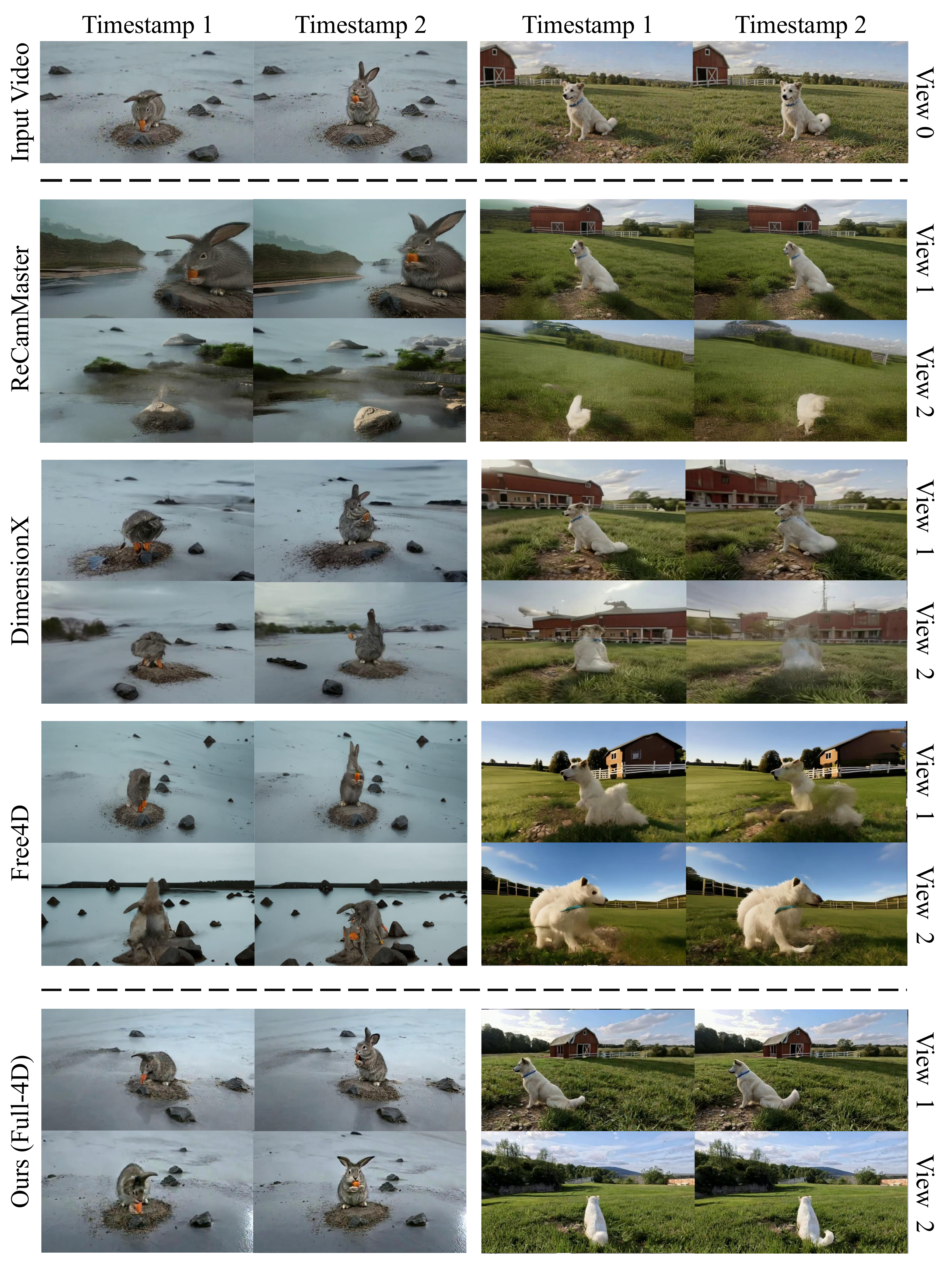}
    \caption{\textbf{Additional qualitative comparisons.} This figure presents two representative cases (left and right), where all methods are evaluated using the same input video. For each case, the first and second columns correspond to two different timestamps, and the first and second rows correspond to two significantly different viewpoints.}
    \label{fig:case2}
\end{figure}

\begin{figure}
    \centering
    \includegraphics[width=1\linewidth]{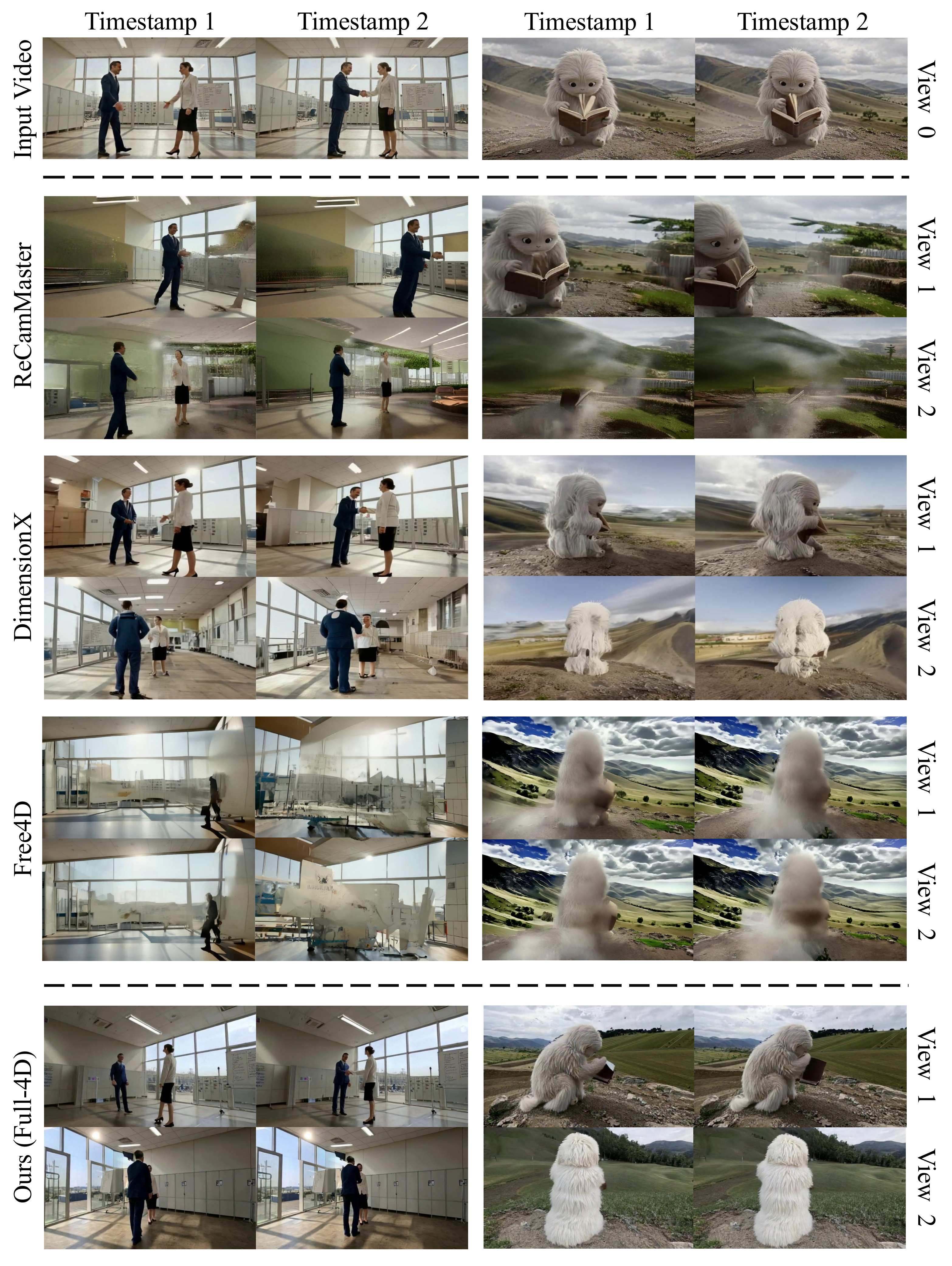}
    \caption{\textbf{Additional qualitative comparisons.} This figure presents two representative cases (left and right), where all methods are evaluated using the same input video. For each case, the first and second columns correspond to two different timestamps, and the first and second rows correspond to two significantly different viewpoints.}
    \label{fig:case3}
\end{figure}

In addition to the quantitative comparisons on our benchmarks, we provide further qualitative evaluations with state-of-arts baselines across diverse data in Fig.\ref{fig:case1},\ref{fig:case2},\ref{fig:case3}. To thoroughly assess the robustness of our method under extreme viewpoint variations, we visualize two significantly different rotations for each scene, ranging from nearly lateral views around ninety degrees to even fully backside perspectives. Experimental results demonstrate that existing baseline methods exhibit clear limitations when handling large-angle rotations: they either fail to generate plausible content beyond a certain rotation angle, or suffer from severe spatial distortions, structural deformations, and content inconsistencies under substantial viewpoint changes. In contrast, our method  achieves smooth and continuous large-range viewpoint transformations while preserving structural integrity and identity consistency, demonstrating superior generative capability and geometric stability.

\begin{figure}
    \centering
    \includegraphics[width=0.9\linewidth]{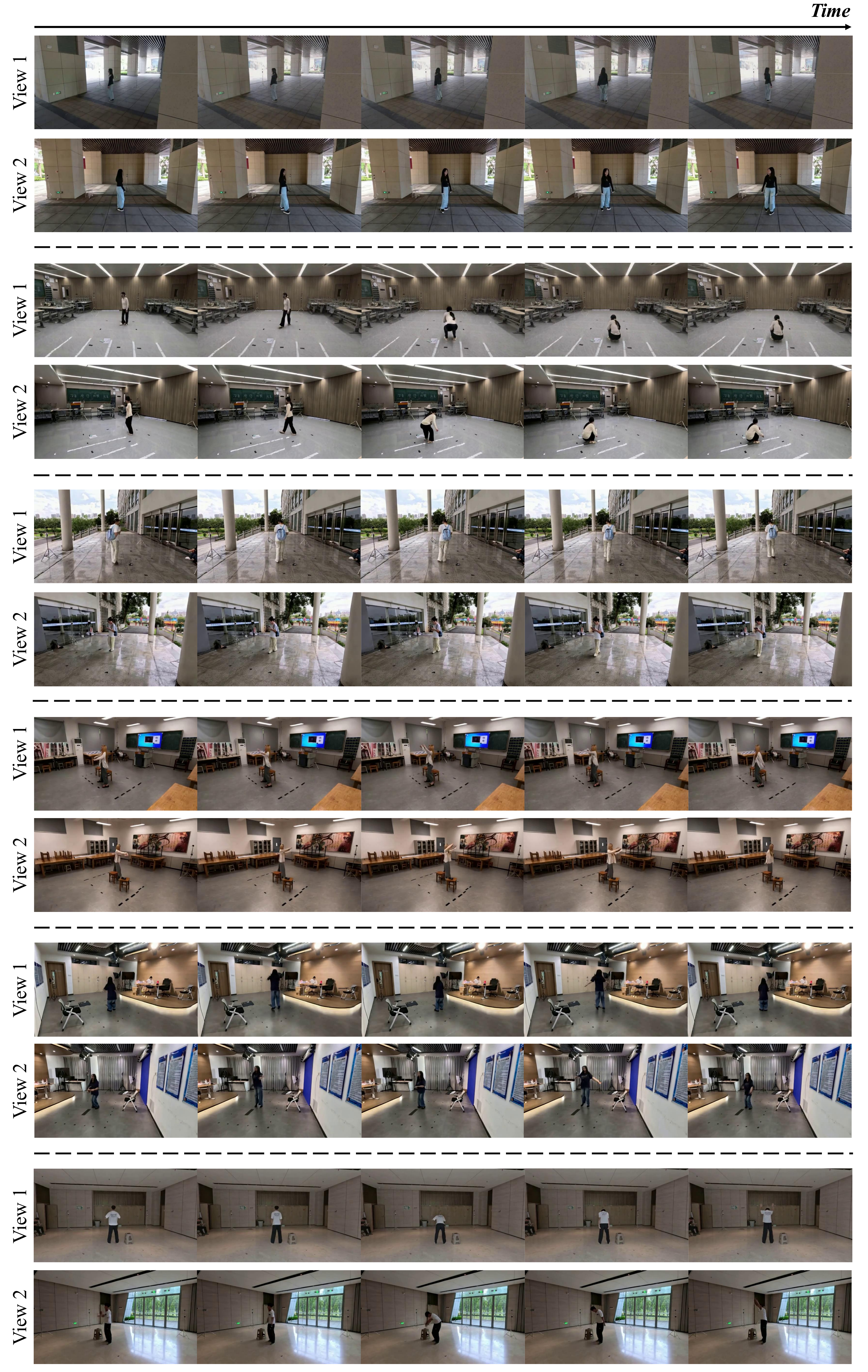}
    \vspace{-2mm}
    \caption{\textbf{Results of our method on diverse test scenes.} Each case (separated by dashed lines) shows two viewpoints (rows), with time progressing from left to right.}
    \label{fig:ours}
\end{figure}

Furthermore, we show results with longer frame sequences across diverse scenes from our test set in Fig.\ref{fig:ours}, demonstrating both temporal continuity and cross-view consistency over extended durations. 

\section{Discussion}
Modeling 4D scenes inherently requires view-time integration. However, existing approaches often treat this integration as a generic feature fusion problem, lacking explicit geometric grounding. Specifically, CAT4D\cite{cat4d} relies on 3D attention, injects time only via embeddings, and uses iterative alternating sampling for dense grids; 4Real-Video-V2\cite{4realvideov2} applies a symmetric sparse mask, lacking explicit camera-pose conditioning and controllable target views.
In contrast, our method creates a routing mechanism that binds camera parameters, geometric reprojection priors, and view-time interactions within attention. This allows arbitrary-view rendering during generation and FMD-based reconstruction by explicitly injecting these physical conditions—strictly aligning generation with physical 3D priors, unlike previous basic feature fusion.

\section{Limitations and Future Work}
To our knowledge, this is the first work to achieve full-scope 4D scene generation from a single video with wide-baseline multi-view coverage. Given the inherent difficulty of this task, some limitations remain. For example, when the input video captures only the back view of a subject, generating plausible frontal facial features becomes extremely challenging due to the lack of visual evidence. Additionally, our real-world training data may cause style shifts on synthetic content, and depth estimation errors can propagate in complex scenes. Currently, generating six synchronized videos takes around 10 minutes on a single GPU, which is comparable to running the base model multiple times sequentially. 
Further engineering effort with parallelization across multiple GPUs is potential to reduce this to match the speed of single-video generation.

Future work will explore incorporating stronger priors to better hallucinate unobserved regions, as well as data augmentation strategies such as style transfer to improve generalization across domains. We also aim to establish comprehensive benchmarks tailored for full-scope 4D evaluation, and integrate more robust depth estimation techniques to further enhance reconstruction stability.


\clearpage
\bibliography{paper}
\bibliographystyle{authordate1}
\end{document}